\documentclass[10pt,journal,compsoc]{IEEEtran}
\ifCLASSOPTIONcompsoc
  \usepackage[nocompress]{cite}
\else
  \usepackage{cite}
\fi
\usepackage{graphicx}
\usepackage{tikz}
\usepackage{comment}
\usepackage{amsmath,amssymb} 
\usepackage{color}

\usepackage{multirow}
\usepackage{xcolor}
\usepackage{booktabs}

\ifCLASSINFOpdf
\else
\fi
\begin{document}
%


\title{BridgeNet: Comprehensive and Effective Feature Interactions via Bridge Feature for Multi-task Dense Predictions}




%
%
%
%

\author{Jingdong~Zhang,~Jiayuan Fan*,~Peng Ye,~Bo Zhang,~Hancheng Ye,~Baopu Li,~Yancheng Cai,~\IEEEmembership{Tao Chen,~Senior~Member,~IEEE}
\IEEEcompsocitemizethanks{
\IEEEcompsocthanksitem Jingdong Zhang is with the School of Information Science and Technology, Fudan University, Shanghai 200433, China, and also with the Department of
Computer Science, Texas A$\&$M University, College Station, TX 77843 USA (e-mail: jdzhang@tamu.edu). \protect\\

\IEEEcompsocthanksitem Jiayuan Fan is with the Academy for Engineering and
Technology, and Peng Ye, Hancheng Ye, Yancheng Cai, Tao Chen are with the School of Information Science and Technology, Fudan University, Shanghai 200433, China (Corresponding author: Jiayuan Fan, E-mail: jyfan@fudan.edu.cn).\protect\\

\IEEEcompsocthanksitem Bo Zhang is with the Shanghai AI Laboratory, Shanghai 200232, China (e-mail: bo.zhangzx@gmail.com). \protect\\

\IEEEcompsocthanksitem Baopu Li is an independent researcher, 100 Oracle Pkwy, Redwood City, CA 94065 (E-mail: Bpli.cuhk@Gmail.com).}
}

%
%


\markboth{IEEE TRANSACTIONS ON Pattern Analysis and Machine Intelligence}{Zhang \MakeLowercase{\textit{et al.}}: BridgeNet: Comprehensive and Effective Feature Interaction via Bridge Feature for Multi-task Dense Predictions}

%



\IEEEtitleabstractindextext{%
\begin{abstract}

Multi-task dense prediction aims at handling multiple pixel-wise prediction tasks within a unified network simultaneously for visual scene understanding. However, cross-task feature interactions of current methods are still suffering from incomplete levels of representations, less discriminative semantics in feature participants, and inefficient pair-wise task interaction processes. To tackle these under-explored issues, we propose a novel BridgeNet framework, which extracts comprehensive and discriminative intermediate Bridge Features, and conducts interactions based on them. Specifically, a Task Pattern Propagation (TPP) module is firstly applied to ensure highly semantic task-specific feature participants are prepared for subsequent interactions, and a Bridge Feature Extractor (BFE) is specially designed to selectively integrate both high-level and low-level representations to generate the comprehensive bridge features. Then, instead of conducting heavy pair-wise cross-task interactions, a Task-Feature Refiner (TFR) is developed to efficiently take guidance from bridge features and form final task predictions. To the best of our knowledge, this is the first work considering the completeness and quality of feature participants in cross-task interactions. Extensive experiments are conducted on NYUD-v2, Cityscapes and PASCAL Context benchmarks, and the superior performance shows the proposed architecture is effective and powerful in promoting different dense prediction tasks simultaneously.

\end{abstract}

\begin{IEEEkeywords}
Multi-task Learning, Dense Prediction, Representation Learning.
\end{IEEEkeywords}}

\maketitle

\IEEEdisplaynontitleabstractindextext

%
\IEEEpeerreviewmaketitle

\IEEEraisesectionheading{\section{Introduction}\label{sec:intro}}

%
%
%
%
\IEEEPARstart{D}{ense} prediction tasks that predict the pixel-wise label for an image have received much attention in many fields such as self-driving~\cite{li2022bevformer,liu2023petrv2}, surveillance~\cite{cai2023rethinking,lan2018pedestrian}, \textit{etc}. With the aid of Convolution Neural Networks (CNNs), a large number of dense prediction works have made great progress recently, including pose estimation~\cite{wei2016convolutional,sun2019deep,newell2016stacked,chen2018cascaded}, semantic segmentation~\cite{long2015fully,chen2017deeplab,badrinarayanan2017segnet,zhao2017pyramid,ronneberger2015u,liang2020video}, and depth estimation~\cite{eigen2014depth,laina2016deeper,yang2020d3vo,eigen2015predicting}. However, these works mainly focus on one specific task and the high relevance between different dense prediction tasks is under-explored. 

Multi-task Learning (MTL) aims to learn a model that handles multiple different tasks simultaneously. Recently, deep learning-based MTL methods~\cite{gao2019nddr,liu2019end,vandenhende2020mti,zhang2019pattern,heuer2021multitask, ye2022inverted} have achieved great progress in multiple dense prediction tasks. By sharing parameters of the heavy image encoders, MTL methods can also achieve joint training and inference of different tasks with high efficiency. Besides, the high relevance among different dense prediction tasks can be further explored by designing a task-interactive module, which leads to mutual boosting of multi-task performance. 

\begin{figure*}[t]
  \centering
  \includegraphics[width=1.0\linewidth]{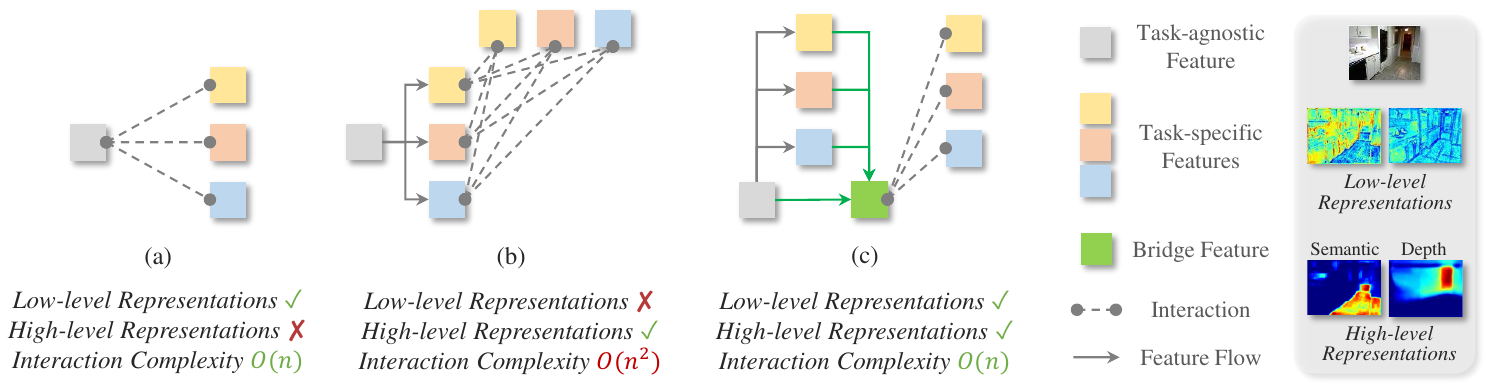}
  \caption{Illustration of multi-task interaction strategies. (a) Encoder-focused~\cite{misra2016cross,liu2019end,gao2019nddr,lu2017fully,vandenhende2019branched,guo2020learning,bruggemann2020automated}: the task-specific features directly interact with the task-generic features from the common backbone. The task-generic features contain rich low-level representations but lack high-level representations. The interaction complexity is $O\left(n\right)$. (b) Decoder-focused~\cite{xu2018pad,crawshaw2020multi,bruggemann2021exploring,zhang2019pattern,zhou2020pattern,ye2022inverted}: the task-specific features are firstly produced by deep supervision, and then interactions are conducted only based on task-specific features, which has discriminative high-level representations, but low-level representations are absent. The pair-wise interaction complexity is $O\left(n^{2}\right)$. (c) BridgeNet (ours): the bridge features are produced from both task-generic and task-specific features and have both rich high-level and low-level representations. The subsequent interactions based on bridge features only have $O\left(n\right)$ complexity. On the right part of the legend, we show some examples of representations from different levels in the feature map: the \textbf{low-level representations} contain less discriminative task information but are rich in details like boundaries, corners and textures, on the contrary, the \textbf{high-level representations} are smooth with less image details, but highly correlated to the task properties, like the highlighted floor with semantic, or areas with larger distance from the depth sensor.}
  \label{fig:fig1}
\end{figure*}

According to where task interactions happen, existing multi-task works are roughly divided into encoder-focused and decoder-focused methods~\cite{vandenhende2021multi}. In particular, the encoder-focused methods conduct interactions of different tasks in the encoding stage, as shown in Fig.~\ref{fig:fig1} (a), task-specific features interact directly with task-generic features produced by the shared backbone, to gain knowledge from multi-task shared representations and thus achieve indirect cross-task interactions. Since the image backbone/encoder is usually initialized with pre-trained parameters that contain abundant visual prior knowledge, and are supervised by gradients from all tasks during the training, it can produce stronger representation with rich \textbf{low-level representations}, i.e. fundamental image details such as edges, corners, textures, etc., as shown in the right legend of Fig.~\ref{fig:fig1}. For pixel-level dense prediction tasks like semantic segmentation, these representations are essential in gaining basic pixel relations and forming accurate semantic masks. However, for joint dense predictions among multiple tasks, these low-level representations are task-generic, which means they only contain general image structures, but lack distinguishable semantics that relate to each specific task. Directly conducting interactions based on task-generic features is sub-optimal, since the mining of discriminative task-specific features is insufficient and consequently limits the role of cross-task interaction~\cite{vandenhende2021multi}.

To alleviate the above issue, the decoder-focused methods conducting interactions in the decoding stage have been developed, where discriminative \textbf{high-level representations}, i.e. abstract and semantic representations derived from images such as key-points, objects, geometries, etc., is perceived from ground-truth labels to facilitate feature interactions, as shown in the right legend of Fig.~\ref{fig:fig1}. Specifically, the initial task predictions are firstly produced by preliminary decoders with deep supervision in the early decoding stage~\cite{xu2018pad,zhang2019pattern,zhou2020pattern,vandenhende2020mti,ye2022inverted}, aiming to decouple the generic encoder features and discover task-specific high-level representations according to the corresponding labels. Subsequently, the cross-task interactions are conducted based on high-level representations by modules such as multi-modal distillation~\cite{xu2018pad} or the ATRC~\cite{bruggemann2021exploring}. These methods try to model task-pair relations, as shown in Fig.~\ref{fig:fig1} (b), by extracting and transferring useful task knowledge from one task to another based on task-specific features, most of them achieve better performance compared with encoder-focused methods. 

Despite the promising performance improvement, these previous methods still face challenges:

\noindent i) Firstly, the cross-task interactions are suffering from incomplete levels of representations in the participants, i.e. either task-generic features with low-level representations or task-specific features with high-level representations are exploited in prior works, which leads to incomplete participants for feature interaction. Though decoder-focused methods have validated that the interactions with discriminative task-specific features are more effective, in which high-level representations are gained in the initial decoding stage. However, the essential low-level representations with abundant basic image details gradually disappear since they are not directly required by supervision signals, and subsequently, in the following task interaction stage, only high-level representations are involved while low-level representations that produce essential dense prediction priors among different tasks are absent, leading to incomplete task interactions and limiting the model performance. 

\noindent ii) Secondly, the cross-task interactions are also suffering from less discriminative semantics in the participants. We observe that the task-specific features produced by the preliminary decoders are usually of low quality with less discriminative representations (as shown in Fig~\ref{fig:vis-attn} right), which negatively affects the subsequent interactions based on them. By analyzing the \textbf{task patterns}, which are defined as the attention distributions or feature structures specifically related to a certain task (as shown in Fig.~\ref{fig:vis-attn} left), we discovered that task patterns significantly differ on different tasks. While in the task-generic features produced by the shared encoder, patterns are usually highly entangled, which makes it difficult to decompose each specific task pattern, and simply imposing supervision on preliminary decoders is not able to fundamentally solve this task-pattern entanglement issue. Thus a more effective way of disentangling meaningful and semantically high-level representations from the shared task-generic features is required for producing high-quality interaction participants.

\noindent iii) The cross-task interaction way is inefficient with high extension costs. Here exemplified by Fig. \ref{fig:fig1} (b). Since previous decoder-focused models need to consider pair-wise task relations~\cite{xu2018pad,vandenhende2020mti,bruggemann2021exploring,ye2022inverted}, with the task number increasing, the complexity of task interactions will increase in $O\left(n^{2}\right)$, which is costly and limits the extension of previous methods on the settings with more tasks.

\begin{figure*}[t]
  \centering
  \includegraphics[width=1.0\linewidth]{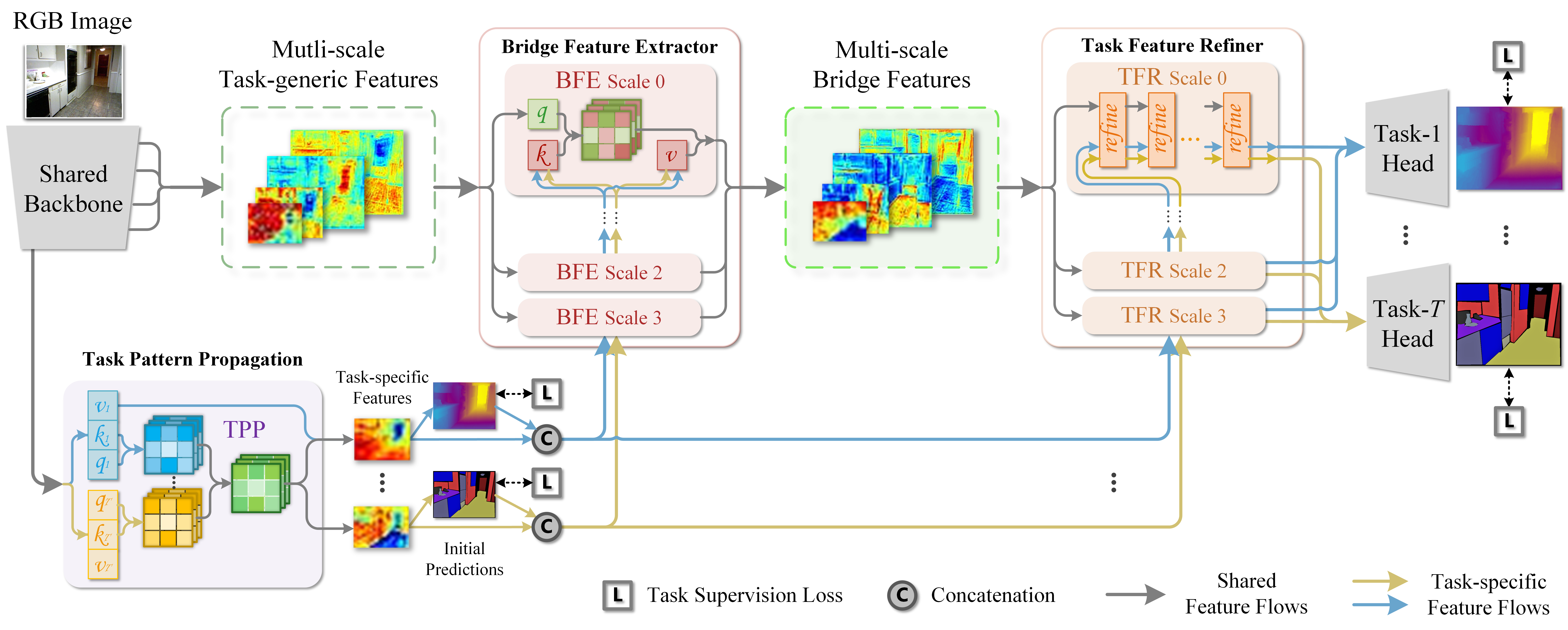}
  \caption{The overview of our proposed method. We take depth estimation and semantic segmentation as examples. 
  We use a shared image backbone to encode task-generic features, and a set of preliminary decoders with deep supervision are used for task-specific feature extraction. Different from decoder-focused methods, a TPP is added before the initial predictions are formed to tackle the task-pattern-entanglement issue. The produced high-quality task-specific features are not used for interaction directly, they are firstly processed by BFE at each scale to gain low-level representations and form the bridge features. Then, the multi-scale bridge features are used to gradually refine task-specific features by TFR. The outputs of TFR are aggregated and fed into task-specific heads for final predictions.
    The detailed structures of TPP (Sec. \ref{sec:S-MSA}, Fig. \ref{fig:s-msa} (a)), BFE (Sec. \ref{sec:DS}, Fig. \ref{fig:BFE}) and TFR (Sec. \ref{sec:CMSAFPN}, Fig. \ref{fig:s-msa} (b)) will be given later.}
  \label{fig:main}
\end{figure*}

To this end, we focus on improving the interaction quality for superior Multi-task Learning from the above three aspects, and propose a novel BridgeNet framework, which is based on the interaction with a comprehensive intermediate feature, namely the \textbf{Bridge Feature}. As shown in Fig.~\ref{fig:fig1} (c), to solve the first challenge, we propose to extract and take advantage of the bridge feature which contains both rich low-level and high-level representations to ensure the integrity of feature interactions. To solve the second challenge, we propose to disentangle task patterns by jointly learning and distributing task patterns to corresponding tasks, in order to produce high-quality task-specific features serving as interaction participants. And to solve the third challenge, we propose to conduct interactions directly between the bridge features and each task-specific feature, which only involves $O\left(n\right)$ complexity of task interactions. Specifically, our proposed method consists of a shared encoder backbone for the task-generic features production, the preliminary decoders with Task Pattern Propagation (TPP) in the early decoding stage producing high-quality task-specific features and handling the task-pattern-entanglement issue. The specially designed Bridge Feature Extractor (BFE), which has a transformer-based structure with cross-attention. By globally querying task-specific features with the task-generic features at each scale, the BFE selects high-level representations correlated to every task and injects them into the task-generic features with rich low-level representations. Thus the extracted powerful bridge features contain comprehensive task representations in both high-level and low-level, and fertilize the formation of final predictions by the Task Feature Refiner (TFR). The overall bridge-feature-based multi-task framework is named as BridgeNet. 

The main contributions of our work are three-fold:
\begin{itemize}
    \item[1)] We discover that incomplete, low-quality interaction participants and inefficient interaction processes exist in current cross-task interaction designs of multi-task dense prediction frameworks. To tackle these issues, we propose the novel BridgeNet framework, which performs a comprehensive interaction based on the bridge features containing both low-level and high-level representations. To the best of our knowledge, bridge features with comprehensive types of representations are involved in the multi-task dense prediction for the first time, which are exploited to transfer cross-task knowledge and fertilize the final task predictions.
    
    \item[2)] A Task Pattern Propagation (TPP) module is firstly applied to avoid the task-pattern-entanglement issue and prepare high-quality participants for the subsequent interactions. A transformer-based Bridge Feature Extractor (BFE) is designed to extract bridge features from both task-generic and task-specific features for comprehensive interactions. Finally, a Task Feature Refiner (TFR) is applied to take advantage of bridge features and refine the final task predictions.

    \item[3)] Our proposed method is extensively evaluated on various dense prediction tasks, including semantic segmentation, depth estimation, surface normal estimation, saliency estimation, and edge detection. The experimental results and insightful analyses on NYUD-v2 and PASCAL Context datasets demonstrate that the proposed architecture achieves superior performance over the state-of-the-art works.
\end{itemize}


\begin{figure*}[t]
  \centering
  \includegraphics[width=0.85\linewidth]{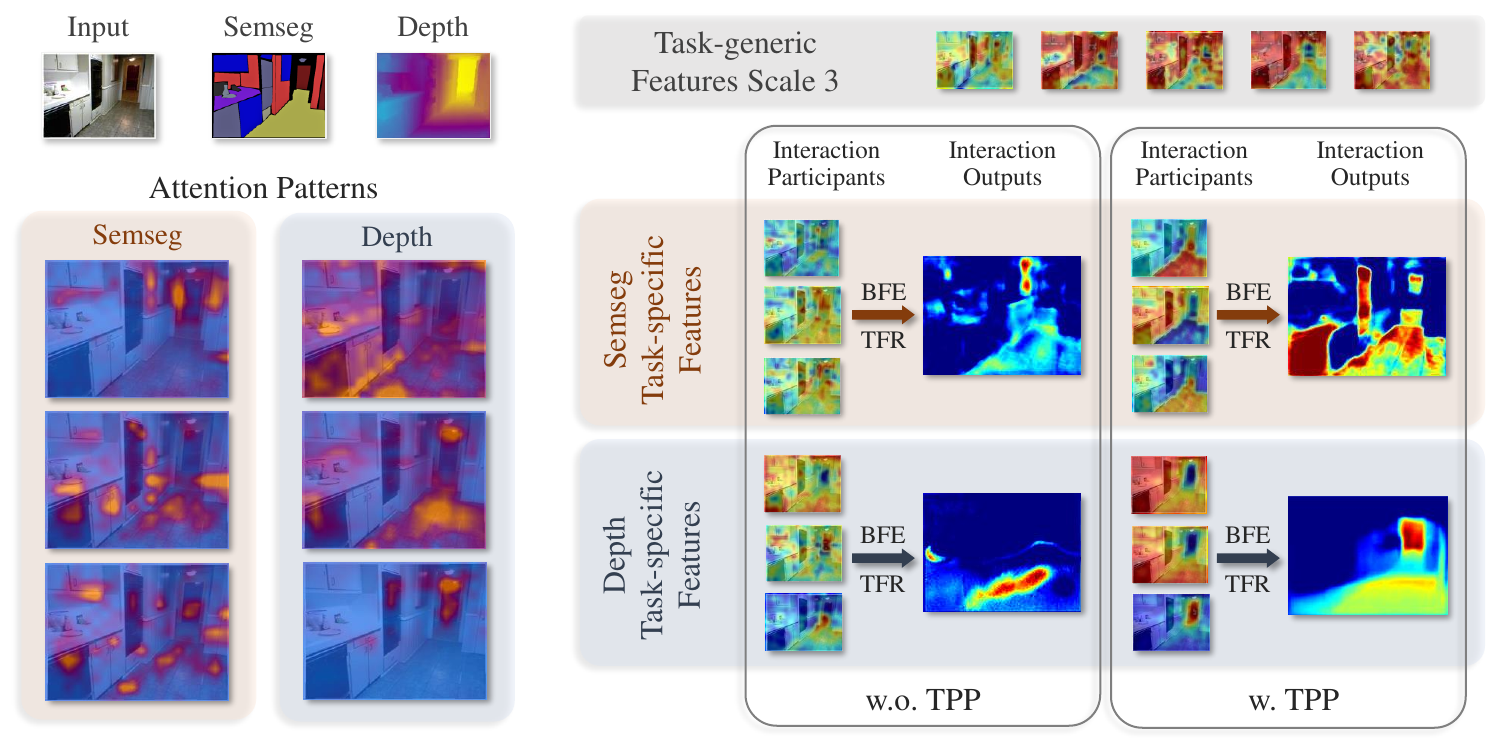}
  \caption{\textbf{Left:} Visualization of different patterns of two tasks (semantic segmentation and depth estimation). The semantic feature focuses on objects with various semantic information, like cabinets and floors, while depth attention focuses more on the surface and edge features with geometry information. \textbf{Right:} Visualization of the task-specific features with (w.) and without (w.o.) TPP. The task-generic features produced by the shared encoder show the task-pattern-entanglement issue, which is significantly different from the distributions of task labels, and the contained patterns are implicit and ambiguous. Without TPP, the decomposed task-specific features serving as the interaction participants are still struggling with the lack of discriminative semantics, which negatively affects the subsequent interaction process (BFE, TFR) and eventually produces low-quality interaction outputs. However, with TPP, the task-specific features can obtain well-decomposed representations that have more similar distributions to their ground-truth labels (like the highlighted floor area in semantic segmentation and door area in depth estimation), and thus boosts the subsequent interaction process to produce high-quality discriminative features. }
  \label{fig:vis-attn}
\end{figure*}

\section{Related Works}\label{sec:formatting}

\noindent\textbf{Multi-Task Learning Architectures: }
With the rapid development of deep CNNs, a lot of MTL works have achieved promising results~\cite{misra2016cross,gao2019nddr,liu2019end,xu2018pad,zhang2018joint,zhang2019pattern,zhou2020pattern,lu2017fully,vandenhende2019branched,vandenhende2020mti,bruggemann2020automated,ruder2019latent,guo2020learning,li2020knowledge,ye2019student,wu2019mutual,heuer2021multitask,ye2022inverted}. As mentioned in Sec. \ref{sec:intro}, existing MTL works are roughly divided into encoder-focused and decoder-focused architectures. In addition, the sharing ways of network parameters are categorized into hard and soft parameter sharing~\cite{vandenhende2021multi,crawshaw2020multi}, which correspond to directly shared parameters or indirectly constrained by regularization or losses, respectively. The encoder-focused models in~\cite{gao2019nddr,ruder2019latent,misra2016cross} have specifically designed shared structures to explore backbone task-generic features in the encoding stage, while the decoder-focused models in~\cite{vandenhende2021multi} adopt hard parameters sharing in the encoding stage to reduce redundant parameters and FLOPs, and specially focus on the interaction among different task-specific features in decoding stages. In this paper, our proposed method is based on the decoder-focused architecture, however, also takes advantage of the encoder-focused models, i.e. the exploring of task-generic features with rich low-level representations, and computational-friendly multi-task interaction ways.



\noindent\textbf{Task Interaction Strategies in MTL: }
In MTL, the learned representation from one task may be beneficial for other tasks, so it is essential to design task interaction manners to promote mutual performances, and avoid potential information inconsistencies. In the field of knowledge distillation, a multi-task student network is distilled through several single-task teacher networks, which transfer multi-task knowledge to the student network and achieve good performance~\cite{ye2019student,li2020knowledge}. However, this network requires to pretrain a group of teacher networks which costs a lot by increasing the number of tasks. The typical encoder-focused models~\cite{misra2016cross,gao2019nddr,liu2019end} conduct interactions at the encoding stage by sharing parts of parameter spaces, especially,~\cite{liu2019end} designs attention modules for feature extraction. While another group of works further studies the parameters sharing scheme in the encoding stage and develops the Branched MTL~\cite{lu2017fully,vandenhende2019branched,guo2020learning,bruggemann2020automated}, aiming at manually or automatically determining the task-shared and task-specific branches to minimizing task inconsistencies. In contrast, decoder-focused models, such as PAD-Net~\cite{xu2018pad}, employ the multi-model distillation based on initial predictions by deep supervision at the decoding stage, where features of auxiliary tasks are extracted and transferred to target tasks to improve their performance. MTI-Net~\cite{vandenhende2020mti} argues that the information of different tasks only promotes each other at certain scales, and extends the multi-model distillation to the multi-scale fashion. PAP-Net~\cite{zhang2019pattern} and PSD~\cite{zhou2020pattern} apply the affinity and the interactions of tasks at pixel and pattern-structure levels respectively. ATRC~\cite{bruggemann2021exploring} designs automatically searching for source-to-target task relations by adaptive task-relational context heads with Neural Architecture Search (NAS) techniques. InvPT~\cite{ye2022inverted} is the first transformer-based joint learning architecture of global spatial interaction and simultaneous all-task interaction. Although decoder-focused models have recently developed rapidly with a large number of variants, and achieve significant improvements, the interaction of them is only based on task-specific features. Besides, the idea of directly source-to-target task-pair relation modeling is inefficient. To tackle the information loss and high computational cost of these methods, we adopt a brand new BridgeNet architecture, which absorbs the advantages of both encoder- and decoder-focused methods, in order to maintain comprehensive task interaction along with acceptable resource cost.

\begin{figure*}[t]
  \centering
  \includegraphics[width=1.0\linewidth]{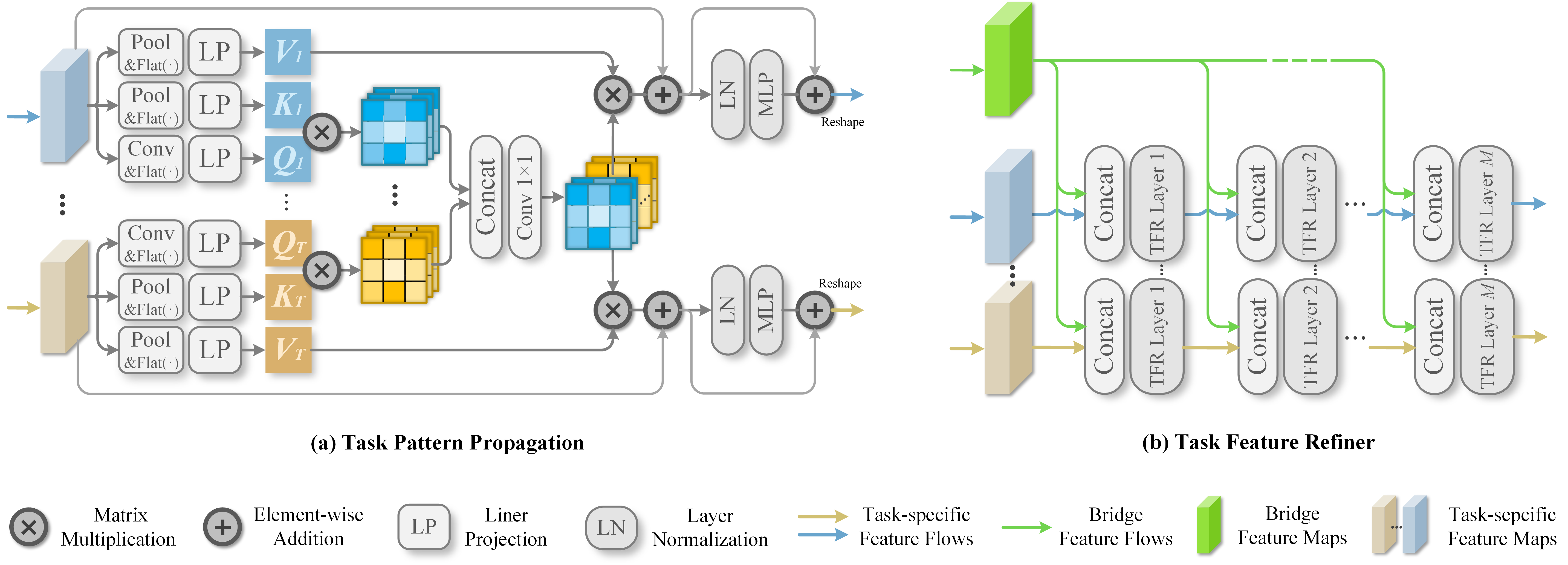}
  \caption{(a) The structure of TPP which is applied after the last backbone layer. The propagation attention map shares the patterns for each task. (b) Task Feature Refiner (TFR), employs a cascaded structure that can be flexibly deployed within the decoding process to inject rich representations from bridge features into each task-specific feature. TFRs of different scales have similar structures and we illustrate the specific structure of one particular scale here.}
  \label{fig:s-msa}
\end{figure*}

\noindent\textbf{Attention in MTL: }
Recently, many MTL architectures~\cite{liu2019end,xu2018pad,zhang2019pattern,vandenhende2020mti,bruggemann2021exploring,ye2022inverted,zhang2024multitask} have employed attention modules to dynamically select information among different tasks. In~\cite{liu2019end}, attention is used to extract features of different tasks from the shared backbone.~\cite{xu2018pad,vandenhende2020mti} distill auxiliary tasks through the attention to extract useful features for target tasks. Since self-attention is capable of capturing global receptive fields by considering the global correlation in feature maps~\cite{wang2018non,fu2019dual}, some MTL works directly introduce self-attention into networks~\cite{zhang2019pattern}. However, these works do not further explore the distribution and interaction of attention among different tasks. With the development of Transformer~\cite{vaswani2017attention} in vision~\cite{dosovitskiy2020image,wang2021pyramid,wu2021cvt,liu2021swin,xie2021segformer,carion2020end}, Multi-head self-attention (MHSA) becomes popular for processing visual scenarios due to its long-range dependencies and context-capture capability. We further study the patterns of MHSA among different tasks and propose the Task Pattern Propagation (TPP) which produces discriminative task-specific features in the early decoding stage to solve the task-pattern-entanglement issue. The transformer-based BFE produces bridge features by global cross-attention between task-generic and task-specific features. Different from simply applying attention to MTL architectures, we focus on structural analysis of task features in the whole interaction process, and introduce priors (e.g. low-level image representations, task patterns) with special design accordingly.

\section{The Proposed Method}
\subsection{Overview}
The overview of the proposed BridgeNet is shown in Fig.\ref{fig:main}, which mainly consists of TPP for task-pattern disentanglement, BFE for bridge feature extraction, and TFR for task-specific feature refinement. In the early decoding stage, we gain multi-scale task-generic features $\mathbf{S}=\left\{S_{i} \mid i=0,1,2,3\right\}$ from the shared image backbone, which can be ConvNets (e.g. HRNet, ResNet) or Vision Transformers (e.g. ViT). The task-generic features are firstly partitioned into patch tokens and embedded into the decoding dimensions. Simultaneously, a set of preliminary decoders with deep supervision are implemented to produce task-specific features $\mathbf{P}=\left\{P^{j}_{i} \mid i=0,1,2,3 ;\quad j=0,1,\cdots,T\right\}$, where $T$ represents the total number of tasks. The TPP is applied before the feature is fed into initial prediction heads to avoid task-pattern entanglement. Subsequently, both task-generic and -specific features are fed into BFE to produce the multi-scale bridge features $\mathbf{S^{\prime}}=\left\{S^{\prime}_{i} \mid i=0,1,2,3\right\}$. After the bridge features are formed, they are in turn used to fertilize task-specific features by TFR which transfers helpful representations from bridge features to task-specific features. Both BFE and TFR are applied on multiple scales, the task-specific features are up-sampled by shared up-sampling layers at each scale. Finally, the refined multi-scale task-specific features are aggregated and fed into prediction heads to make final predictions. The detailed structures will be described in the next sections.

\begin{figure*}[t]
  \centering
  \includegraphics[width=1.0\linewidth]{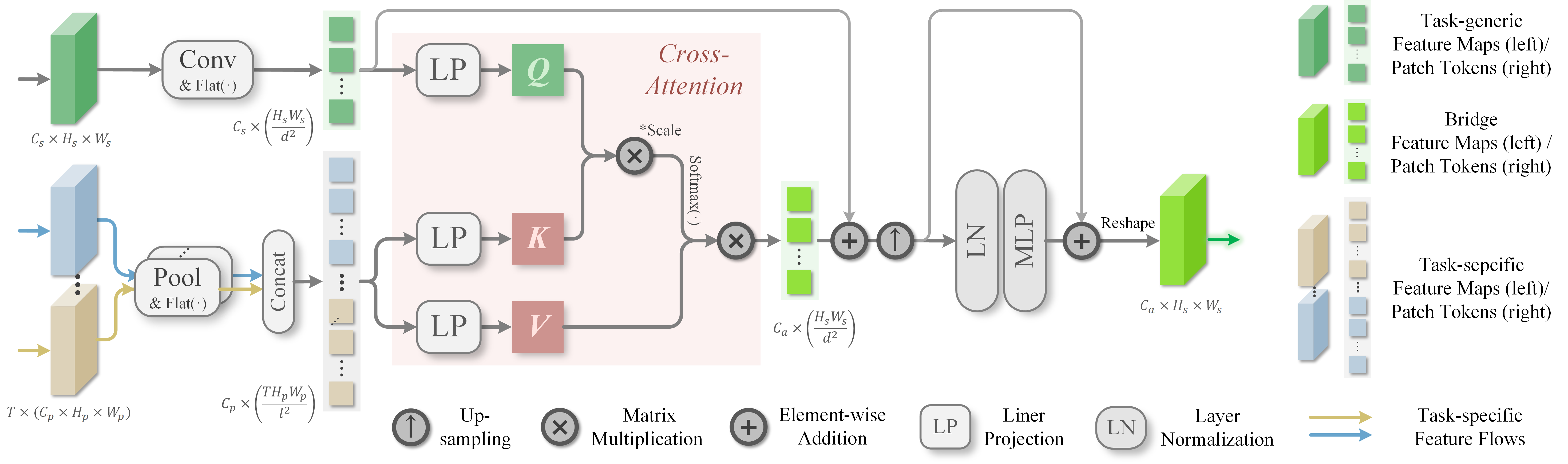}
  \caption{A zoom-in version of one-scale BFE in Fig.2 (Different scales have similar structures). The input task-generic and task-specific features are firstly transformed into patch tokens, and then we use the task-generic tokens to query all of the task-specific tokens globally to produce a global correlation map, which helps select useful high-level information to gain high-level representations.}
  \label{fig:BFE}
\end{figure*}

\subsection{Task Pattern Propagation}\label{sec:S-MSA}
For multi-scale task-generic features, features with smaller scales contain richer high-level semantic and more contextual information and usually guide the formation of predictions in a top-down way. Meanwhile, different dense prediction tasks usually share a lot of similarities in low-level representations, but significantly differ in high-level representations. Thus, for the task-generic features produced by the final layers of the shared backbone, the high-level representations from different tasks are usually entangled severely, making it difficult to extract discriminative representations for each task and subsequently conduct interactions. We name this phenomenon the task-pattern-entanglement issue. As shown in Figure~\ref{fig:vis-attn} (left), we visualize the regions with soft attention scores on semantic segmentation and depth estimation respectively, which show great differences that semantic attention focuses more on different objects and contexts while depth attention focuses more on spatial and geometry information. In Figure~\ref{fig:vis-attn} (right), the patterns in the encoder shared outputs are implicit and entangled, which do not clearly reflect the distributions of either task. Thus, task-specific features directly decomposed from them are struggling with the lack of discriminative semantics, which leads to low-quality interaction participants for the following interactions and eventually brings negative effects to interaction outputs.

To tackle the task-pattern-entanglement issue, we propose to conduct Task Pattern Propagation (TPP) that jointly learns and propagates different task patterns to corresponding tasks to assist in the formation of task-specific features. Although learning and propagating task patterns have been discussed in previous multi-task learning works~\cite{zhang2019pattern,zhou2020pattern}, the entanglement issues at the pattern level remain unexplored and unsolved, and our proposed TPP first time targeting this issue from the task pattern level. As shown in Fig. \ref{fig:s-msa}, firstly, for every task-specific feature maps ${P_{3}^{j}}$ at scale $3$, where $j=1,2, \cdots, T$, {we correspondingly generate $T$ sets of \textit{query} $Q_{j}$, \textit{key} $K_{j}$, and \textit{value} $V_{j}$ by linear projection for each task. Then task-specific attentions $A_{j}, j=1,2, \cdots, T$ are calculated by dot-product accordingly, which contains attention patterns of each task as follows:}

\begin{equation}
A_{j}=\frac{Q_{j} \times K_{j}^{\top}}{\sqrt{C_{a}}},
  \label{eq:n5}
\end{equation}

\noindent where $C_{a}$ denotes the dimension for attention space. Subsequently, attention maps from all tasks $A_{j}$ are concatenated and then squeezed by $1 \times 1$ convolution in a common space to align the dimension and share task patterns. Later the task patterns in attention maps are propagated by dot-product with all of the task \textit{value}s:
\begin{equation}
\begin{array}{c}
\bar{A} = \operatorname{Conv_{1 \times 1}}\left(\operatorname{Concat}\left(A_{1}, A_{2}, \cdots, A_{T}\right)\right),\\
x_{j} = {\operatorname{Softmax_{dim=-1}}}(\bar{A}) \times V_{j}, \quad j=1,2, \cdots, T .
\end{array}
  \label{eq:n6}
\end{equation}

\noindent {where $\operatorname{Softmax_{dim=-1}(\cdot)}$} represents the softmax function applied to the last feature dimension.

Finally, $x_{j}$ are processed by an FFN which has a similar structure in BFE to enhance the learned task-specific high-level representations. Since different tasks have various pattern distributions, it is necessary to perform a pattern propagation to share the attention space for multiple tasks. Empirically, the irrelevant pixels will be repressed during the matching to avoid the negative feature from transferring, which avoids the unexpected entanglement caused by task inconsistencies. With TPP, we extract explicit task patterns by self-attention and conduct propagation to avoid potential entanglements and achieve a clear decoupling effect of task-specific features, which also benefits the subsequent feature interactions of BFE and TFR. As shown in Fig. \ref{fig:vis-attn} {(right)}, TPP can significantly strengthen the task patterns by producing more discriminative high-level semantics, like the highlighted floor area in semantic segmentation and door area in depth estimation, which boosts the subsequent interactions by providing high-quality participants.

\subsection{Bridge Feature Extraction} 
\label{sec:DS}
In this subsection, we discuss the detailed structure of our key component, i.e. the Bridge Feature Extractor (BFE). The BFE is a transformer-based module, that aims at extracting useful high-level representations from all task-specific features and producing bridge features. The core part of BFE is the global modeling of correlations between task-generic features and all of the task-specific features, which selects important high-level information and transfers it to the generic features. The BFE is consecutively stacked at each scale to produce multi-scale bridge features, and the BFE block at each scale has the same structure. As illustrated in Fig~\ref{fig:BFE}, assuming that the input task-generic and -specific features are at scale $i$, i.e. $S_{i} \in \mathbb{R}^{C_{s} \times H_{s} \times W_{s}}$ and $P^{j}_{i} \in \mathbb{R}^{C_{p} \times H_{p} \times W_{p}}, j=0,1,\cdots,T$, they are firstly transformed into patch tokens by depth-wise convolution or average pooling with $stride=d$ and $stride=l$ respectively:

\begin{equation}
\begin{array}{c}
\widetilde{S_{i}}=\operatorname{Flat}(\operatorname{Conv}(S_{i})),\\
\widetilde{P_{i}^{j}}=\operatorname{Flat}(\operatorname{Pool}(P_{i}^{j})), \quad j=0,1,\cdots,T ,
\end{array}
  \label{eq:n1}
\end{equation}

\noindent where $\widetilde{S_{i}} \in \mathbb{R}^{C_{s} \times \left(\frac{H_{s} W_{s}}{{d}^{2}}\right)}$ and $\widetilde{P_{i}^{j}} \in \mathbb{R}^{C_{p} \times \left(\frac{H_{p} W_{p}}{{l}^{2}}\right)}$, $\operatorname{Flat}(\cdot)$ represents the flatten operation for tensors in the spatial dimensions. The patch tokens will receive a spatial reduction which controls the computation cost of the following attention calculation. Afterward, all of the task-specific patch tokens are concatenated at the spatial dimension. To conduct the cross-attention, the \textit{query} ($Q$) is projected from the transformed task-generic patch tokens, the \textit{key} ($K$) and \textit{value} ($V$) is projected from the transformed task-specific patch tokens. The production of $Q$, $K$ and $V$ can be described as:

\begin{equation}
\begin{array}{c}
Q=\operatorname{LP}(\widetilde{S_{i}})^{\top},\\
K=\operatorname{LP}(\operatorname{Concat}(\widetilde{P_{i}^{0}},\widetilde{P_{i}^{1}}, \cdots, \widetilde{P_{i}^{T}})^{\top},\\
V=\operatorname{LP}(\operatorname{Concat}(\widetilde{P_{i}^{0}},\widetilde{P_{i}^{1}}, \cdots, \widetilde{P_{i}^{T}})^{\top},
\end{array}
  \label{eq:n2}
\end{equation}

\noindent where $Q \in \mathbb{R}^{\left(\frac{H_{s} W_{s}}{{d}^{2}}\right) \times C_{a}}$ and {$K, V \in \mathbb{R}^{\left(\frac{T H_{p} W_{p}}{{l}^{2}}\right) \times C_{a}}$,} $C_{a}$ is the channel dimension for attention. Then, we conduct a standard cross-attention:

\begin{equation}
A=\frac{Q \times K^{\top}}{\sqrt{C_{a}}}, \quad A \in \mathbb{R}^{\frac{H_{s} W_{s}}{d^{2}} \times \frac{T H_{p} W_{p}}{l^{2}}},
  \label{eq:n3}
\end{equation}

\noindent after the attention map is calculated, the output of cross-attention is:

\begin{equation}
x = {\operatorname{Softmax_{dim=-1}}}(A) \times V, \quad x \in \mathbb{R}^{\left(\frac{H_{s} W_{s}}{{d}^{2}}\right) \times C_{a}},
  \label{eq:n4}
\end{equation}

 Then $x$ is transposed, reshaped, and up-sampled into ${C_{a} \times \left({H_{s} W_{s}}\right)}$ and fed into a feed-forward network which is composed of a layer normalization and an MLP. The final output is reshaped into ${C_{a} \times H_{s} \times W_{s}}$, which is the bridge feature we denote as $S^{\prime}_{i}$ at scale $i$. By querying all of the task-specific features globally, the task-generic feature gains discriminative high-level representations by selecting the task-specific pixels that have the highest response to the task-generic features in $A$, and the low-level representations are simultaneously reserved by residual paths. Thus, the extracted bridge feature meets the need as a medium for multi-task dense prediction. Different from directly conducting pixel-wise addition of task-generic and -specific features, which might cause unexpected conflicts if there is task inconsistency exists in a certain position, the global querying ensures only the highest responding task-specific pixel is selected and avoids the potential task conflicts.

\subsection{Task Feature Refiner}\label{sec:CMSAFPN}
To conduct effective feature fusion between bridge features $\mathbf{S^{\prime}}$ and task-specific features $\mathbf{P}$, in order to take advantage of the rich representations in $\mathbf{S^{\prime}}$, we propose a Task Feature Refiner (TFR), as shown in the right part of Fig. \ref{fig:s-msa} (b). TFR utilizes bridge features to guide task-specific characteristics and achieve effective interaction between tasks. TFR offers a relatively flexible configuration, employing a cascaded structure where each layer consists of TFR layers with the same structures. Within each layer, bridge feature $S^{\prime}_{i}$ from a certain scale $i$ needs to be concatenated with each task-specific feature $P^{j}_{i}, j=0,1,\cdots,T$, followed by fusion through the TFR layer:

\begin{equation}
\begin{array}{c}
x^{(k)}=\operatorname{TFRlayer}^{(k)}\left(\operatorname{Concat}\left(S_{i}^{\prime}, x^{(k-1)}\right)\right),\\ j=0,1, \cdots, T;\quad k=1,2, \cdots, M;\quad x^{(0)} = P_{i}^{j} ,
\end{array}
  \label{eq:n7}
\end{equation}

\noindent where $M$ represents the total number of layers in TFR, stacking these TFR layers forms the entire TFR module. During this process, the task-specific features $P_{i}^{j}$ undergo progressive refinement through these layers, continuously guided by the bridge features, obtaining abundant task representations. This ultimately generates high-quality task-specific features, laying a solid foundation for the generation of final task predictions.

Additionally, our TFR layers can be flexibly configured. In the simplest case, we can use just one convolutional layer to align the concatenated feature channels and generate predictions. However, this is certainly insufficient, as a single convolutional layer lacks the necessary capacity for nonlinear transformations. Alternatively, Transformer layers can be employed for global relationship modeling, such as InvPT~\cite{ye2022inverted}. Yet, experimental findings reveal that the complex quadratic relationship modeling of the Transformer does not significantly improve the quality of task features. This is partly because the scales of the feature maps are not large at this stage, and global modeling doesn't offer significant advantages. Furthermore, spatial local similarity serves as a more crucial prior for dense predictions. Hence, in our method, we utilize depth-wise separable dilated convolutions to form our fundamental layer. We stack dilated convolutions of varying sizes to ensure sufficient receptive fields and avoid the grid effect associated with dilation. Simultaneously, the local connections brought by convolutions are suitable for extracting vital local information from the bridge features to generate the final predictions.

\begin{figure*}[t]
  \centering
  \includegraphics[width=0.8\linewidth]{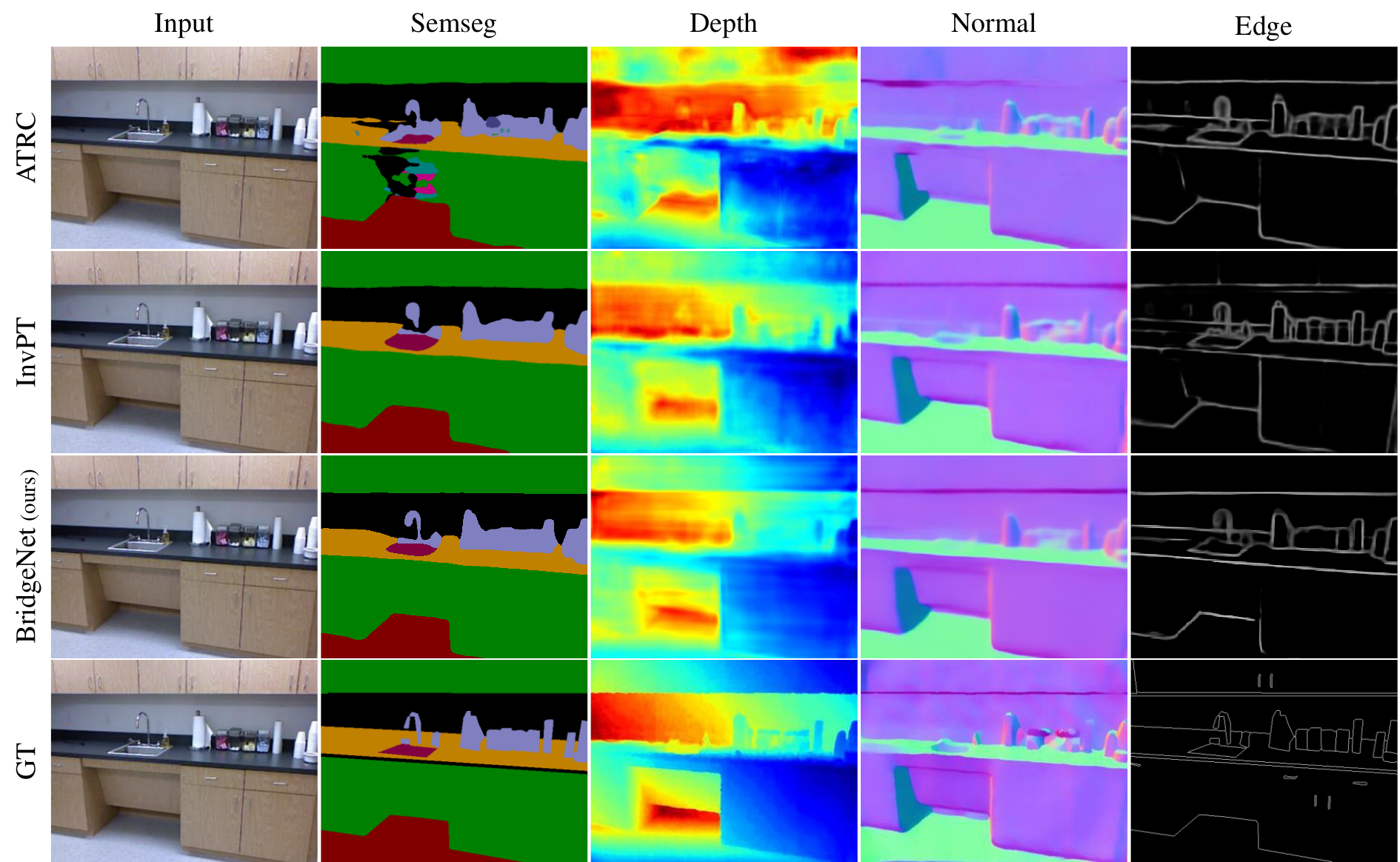}
  \caption{As visualized above, we compare with SOTA works (ATRC\cite{bruggemann2021exploring} and InvPT~\cite{ye2022inverted}) on NYUD-v2 4 tasks. Our BridgeNet clearly produces more precise predictions on each task. To avoid cherry-picking, the input image is selected from the sample examples provided in~\cite{ye2022inverted} paper.}
  \label{fig:com}
\end{figure*}

\section{Experiment}
\subsection{Experimental Setup}
\textbf{Datasets:} We conduct experiments on three benchmark datasets: NYUD-v2~\cite{silberman2012indoor}, Cityscapes~\cite{cordts2016cityscapes} and PASCAL Context~\cite{chen2014detect}. NYUD-v2 dataset is mainly used for indoor scene segmentation and depth estimation in MTL works. The dataset contains 1449 images. Following the standard settings in~\cite{vandenhende2021multi}, we use 795 images for training which are randomly selected, and the rest are for testing. Cityscapes~\cite{cordts2016cityscapes} contains 2975 and 500 high-resolution street-view images captured from different European countries for training and testing respectively. These images are annotated with fine 19-class semantic labels for segmentation and disparity map for depth estimation. To speed up the convergence and train all models in comparisons with controllable computational cost overhead, we down-sample the images with their annotations to $384 \times 768$ resolution. PASCAL is a popular benchmark for many dense prediction tasks and we use the split from PASCAL Context, which contains 10103 images, where 4998 images are randomly selected for training and the rest are for testing~\cite{vandenhende2021multi}. We choose several subsets of tasks on both datasets to make a more explicit comparison, including semantic segmentation (Seg.), human parts segmentation (H.Parts), depth eatimation (Dep.), surface normal estimation (Norm.), saliency estimation (Sal.) and edge detection (Edge.).

\begin{table*}[t]
\caption{Comparison with SOTA works on NYUD-v2 and PASCAL Context.}
\centering
\setlength{\tabcolsep}{2.96mm}{\scalebox{1.0}{
\begin{tabular}{lcccclccccc}
\toprule
\multicolumn{5}{c}{NYUD-v2} & \multicolumn{6}{c}{PASCAL Context}\\ 
\cmidrule(lr){1-5}\cmidrule(lr){6-11}
\multirow{2}{*}{Models}&Seg. &Dep. &Norm. &Edge. & \multirow{2}{*}{Models}& Seg. &H.Parts & Sal. &Norm. &Edge.\\
&\textit{mIoU}↑ &\textit{rmse}↓ &\textit{mErr}↓&\textit{odsF}↑ & &\textit{mIoU}↑ &\textit{mIoU}↑& \textit{maxF}↑ &\textit{mErr}↓ &\textit{odsF}↑\\
\cmidrule(lr){1-1}\cmidrule(lr){2-5}\cmidrule(lr){6-6}\cmidrule(lr){7-11}
Cross-Stitch~\cite{misra2016cross}&36.34	&0.6290	&20.88	&76.38	&ASTMT~\cite{maninis2019attentive}	&68.00	&61.10	&65.70	&14.70	&72.40 \\
PAP~\cite{zhang2019pattern}	&36.72	&0.6178	&20.82	&76.42	&PAD-Net~\cite{xu2018pad}	&53.60	&56.60	&65.80	&15.30	&72.50\\
PSD~\cite{zhou2020pattern}	&36.69	&0.6246	&20.87	&76.42	&MTI-Net~\cite{vandenhende2020mti}	&61.70	&60.18	&84.78	&14.23	&70.80\\
PAD-Net~\cite{xu2018pad}	&36.61	&0.6270	&20.85	&76.38	&ATRC~\cite{bruggemann2021exploring}	&62.69	&59.42	&84.70	&14.20	&70.96\\
MTI-Net~\cite{vandenhende2020mti}	&45.97	&0.5365	&20.27	&77.84	&ATRC-ASPP~\cite{bruggemann2021exploring}	&63.60	&60.23	&83.91	&14.30	&70.86\\
ATRC~\cite{bruggemann2021exploring}	&46.33	&0.5363	&20.18	&77.94	&ATRC-BMTAS~\cite{bruggemann2021exploring}	&67.67	&62.93	&82.29	&14.24	&72.42\\
InvPT~\cite{ye2022inverted}	&53.56	&0.5183	&19.04	&78.10	&InvPT~\cite{ye2022inverted}	&79.03	&67.61	&84.81	&14.15	&73.00\\
BridgeNet (\textit{ours}) &\textbf{56.57}	&\textbf{0.4655}	&\textbf{17.29} &\textbf{80.02} & BridgeNet (\textit{ours}) & \textbf{79.89} &\textbf{71.33} &\textbf{85.64} &\textbf{13.38} &\textbf{73.24} \\
 \bottomrule
\end{tabular}
}}
\label{tab:compare1}
\end{table*}

\begin{table}[t]
\caption{Comparison with SOTA works on Cityscapes.}
\centering
\setlength{\tabcolsep}{8mm}{\scalebox{1.0}{
\begin{tabular}{lcc}
\toprule
\multicolumn{3}{c}{Cityscapes}\\ 
\cmidrule(lr){1-3}
\multirow{2}{*}{Models}&Seg. &Dep.\\
&\textit{mIoU}↑ &\textit{rmse}↓\\
\cmidrule(lr){1-1}\cmidrule(lr){2-3}
Cross-Stitch~\cite{misra2016cross}& 85.31 & 3.422\\
NDDR-CNN~\cite{gao2019nddr}& 84.73 & 3.415\\
MTAN~\cite{liu2019end} & 90.20 & 3.442\\
PAD-Net~\cite{xu2018pad} & 90.13 & 3.496\\
MTI-Net~\cite{vandenhende2020mti} & 91.50 & 2.680\\
InvPT~\cite{ye2022inverted} & 91.19 & 2.631\\
BridgeNet (\textit{ours}) & \textbf{92.61} & \textbf{2.606}\\
 \bottomrule
\end{tabular}
}}
\label{tab:compare1b}
\end{table}

\begin{figure*}[t]
  \centering
  \includegraphics[width=1.0\linewidth]{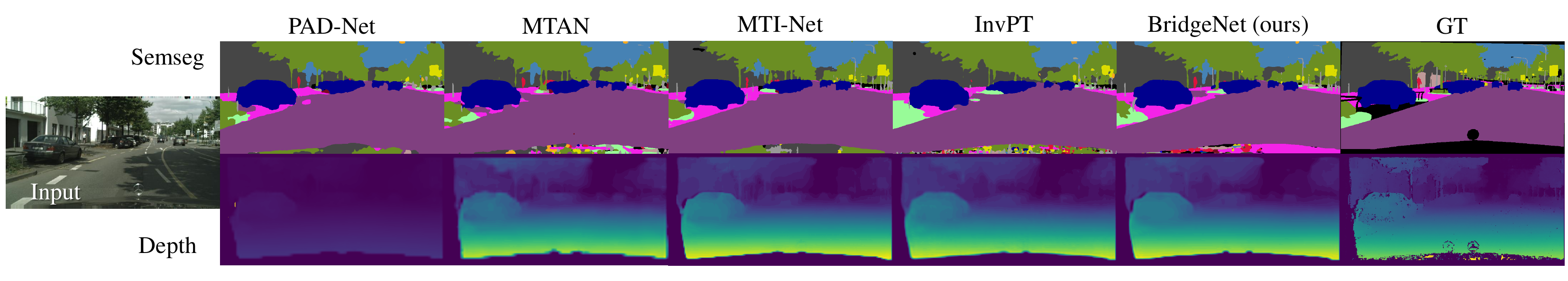}
  \caption{As visualized above, we compare with SOTA works (including PAD-Net~\cite{xu2018pad}, MTAN~\cite{liu2019end}, MTI-Net~\cite{vandenhende2020mti} and InvPT~\cite{ye2022inverted}) on Cityscapes.}
  \label{fig:com_citys}
\end{figure*}

\begin{table*}[t]
\caption{Comparison with SOTA works on different backbones.}
\centering
\setlength{\tabcolsep}{1.78mm}{\scalebox{1.0}{
\begin{tabular}{llccccllccccc}
\toprule
\multicolumn{6}{c}{NYUD-v2} & \multicolumn{7}{c}{PASCAL Context}\\ 
\cmidrule(lr){1-6}\cmidrule(lr){7-13}
\multicolumn{2}{c}{\multirow{2}{*}{Models}}&Seg. &Dep. &Norm. &Edge. & \multicolumn{2}{c}{\multirow{2}{*}{Models}}& Seg. &H.Parts & Sal. &Norm. &Edge.\\
&&\textit{mIoU}↑ &\textit{rmse}↓ &\textit{mErr}↓&\textit{odsF}↑ && &\textit{mIoU}↑ &\textit{mIoU}↑& \textit{maxF}↑ &\textit{mErr}↓ &\textit{odsF}↑\\
\cmidrule(lr){1-2}\cmidrule(lr){3-6}\cmidrule(lr){7-8}\cmidrule(lr){9-13}
\multirow{3}{*}{HRNet-w18} &MTI-Net&	39.89&	0.5824&	20.57&	76.60 &\multirow{6}{*}{ResNet-50d} & ASTMT	&66.80	&61.10	&66.10&	14.70	&70.90 \\
& ATRC	&\textbf{40.80}	&0.5826	&20.51	&76.50& & MTI-Net	&66.60	&\textbf{63.30}	&66.60	&14.60	&\textbf{74.90} \\
& BridgeNet	&40.74	&\textbf{0.5662}	&\textbf{19.95}	&\textbf{77.24}&	&ATRC	&62.69	&59.42	&\textbf{84.70}	&\textbf{14.20}	&70.96 \\
\cmidrule(lr){1-2}\cmidrule(lr){3-6}
\multirow{3}{*}{HRNet-w48}& MTI-Net	&45.97	&0.5365	&20.27	&77.86&	&ATRC-ASPP    &63.60	&60.23	&83.91	&14.30	&70.86 \\
& ATRC	&\textbf{46.33}	&0.5363	&20.18	&77.94&		&ATRC-BMTAS	&67.67	&{62.93}	&82.29	&14.24	&72.42 \\
& BridgeNet	&46.28	&\textbf{0.5292}	&\textbf{19.76}	&\textbf{78.24}&		&BridgeNet	&\textbf{69.79}	&62.49	&84.28	&14.27	&72.30\\ 
\cmidrule(lr){1-2}\cmidrule(lr){3-6}\cmidrule(lr){7-8}\cmidrule(lr){9-13}
\multirow{2}{*}{ViT-base} &InvPT	&50.30	&0.5367	&19.00	&77.86	&\multirow{2}{*}{ViT-base}	&InvPT	&77.33	&66.62	&\textbf{85.14}	&13.78	&\textbf{73.20}\\
&BridgeNet	&\textbf{51.14}	&\textbf{0.5186}	&\textbf{18.92}	&\textbf{77.98}&		&BridgeNet	&\textbf{77.98}	&\textbf{68.19}	&85.06	&\textbf{13.48}	&72.98\\
\cmidrule(lr){1-2}\cmidrule(lr){3-6}\cmidrule(lr){7-8}\cmidrule(lr){9-13}
\multirow{2}{*}{ViT-large}	&InvPT	&53.56	&0.5183	&19.04	&78.10&	\multirow{2}{*}{ViT-large}&	InvPT	&79.03	&67.61	&\textbf{84.81}	&14.15	&\textbf{73.00}\\
& BridgeNet	&\textbf{55.51}	&\textbf{0.4930}	&\textbf{18.47}	&\textbf{78.22}&		&BridgeNet	&\textbf{80.64}	&\textbf{70.06}	&84.64	&\textbf{13.82}	&72.96\\
 \bottomrule
\end{tabular}
}}
\label{tab:compare2}
\end{table*}

\begin{table*}[t]
\caption{Analysis of model sizes and cost. \textbf{Bold} and \underline{underlined} fonts represent the first and second best results respectively.}
\centering
 \setlength{\tabcolsep}{3.2mm}{\scalebox{1.0}{
\begin{tabular}{lccccccccc}
\toprule
\multirow{2}{*}{Models} & FLOPs &Params& Latency& Memory &Seg. &H.Parts & Sal. &Norm. &Edge. \\
& (GB) & (M) & (s) & (GB) &\textit{mIoU}↑ &\textit{mIoU}↑& \textit{maxF}↑ &\textit{mErr}↓ &\textit{odsF}↑\\
\cmidrule(lr){1-1}\cmidrule(lr){2-5}\cmidrule(lr){6-10}
PAD-Net~\cite{xu2018pad} \textit{w.} ViT-large &773 &330 & 0.043 & 3.096 &78.01 &67.12 &79.21 &14.37 &72.60 \\
MTI-Net~\cite{vandenhende2020mti} \textit{w.} ViT-large & 774 & 851 & 0.087 & 7.642 & 78.31 & 67.40 & 84.75 & 14.67 & 73.00 \\
ATRC~\cite{bruggemann2021exploring} \textit{w.} ViT-large & 871 & 340  & 0.098 & 7.933 & 77.11 & 66.84 & 81.20 & 14.23 & 72.10 \\
InvPT~\cite{ye2022inverted} \textit{w.} ViT-large & 669 & 423 & 0.060 & 3.510 & 79.03 & 67.61 & 84.81 &14.15 & 73.00 \\
\cmidrule(lr){1-1}\cmidrule(lr){2-5}\cmidrule(lr){6-10}
BridgeNet \textit{w.} ViT-base &  412 &  230  & 0.090& 2.710& 77.98	&	68.19	&	85.06	&	13.48 & 72.70 \\
BridgeNet \textit{w.} ViT-large & 645 & 477  & 0.144 & 3.718& \textbf{80.64}	&	\underline{70.06}		&84.64		&13.82 & 72.96 \\
\cmidrule(lr){1-1}\cmidrule(lr){2-5}\cmidrule(lr){6-10}
BridgeNet \textit{w.} InternImage-B &  406 & 248  & 0.130 & 2.774 & 78.07 &69.03	&	\underline{85.29}		&\underline{13.44} & \underline{73.10} \\
BridgeNet \textit{w.} InternImage-L &  527& 420  &  0.171 & 3.494 & \underline{79.89} &\textbf{71.33}		&\textbf{85.64}		&\textbf{13.38} & \textbf{73.24} \\
\bottomrule
\end{tabular}
}}
\label{tab:cost}
\end{table*}

\noindent\textbf{Metrics:} We consider multiple metrics for different tasks respectively and conduct extensive experiments to further validate the effectiveness of our model. The metric notations are listed as follows:
\begin{itemize}
\setlength{\itemsep}{0pt}
\setlength{\parsep}{0pt}
\setlength{\parskip}{0pt}
    \item[1)] \emph{mIoU}: mean intersection over union.
    \item[2)] \emph{rmse}: root mean square error. (For surface normal estimation, we calculate the rmse of normal angle.)
    \item[3)] \emph{mErr}: mean of angle error.
    \item[4)] \emph{max-F}: maximum of ${F_{1}}-measure$.
    \item[5)] \emph{odsF}: optimal dataset scale F-measure
\end{itemize}
Additionally, to better evaluate the proposed method, we consider the relative gain in each task, for task ${\tau}_{j}$, $j = 1, \ldots, N$, the relative gain $\Delta_{{\tau}_{j}}$ can be designed as:

\begin{equation}
 \Delta_{{\tau}_{j}}=(-1)^{l_{j}}\left(M_{m, j}-M_{s, j}\right) / M_{s, j},
  \label{eq:13a}
\end{equation}

\noindent where $l_{j}=1$ if a lower value means better for performance measure $M_{j}$ of task $j$, and $l_{j}=0$ if higher is better.

Also, we use the \emph{multi-task performance} $\Delta_{M T L}$ from~\cite{vandenhende2021multi} to evaluate the mutual promotion in all tasks, defined as:

\begin{equation}
 \Delta_{M T L}=\frac{1}{N} \sum_{j=1}^{N}\Delta_{{\tau}_{j}},
  \label{eq:13b}
\end{equation}

\noindent\textbf{Implementation Details:} We conduct our experiments on Pytorch~\cite{paszke2019pytorch} with one NVIDIA Tesla V100 GPU. The models used for different evaluation experiments were trained for $40,000$ iterations on both datasets with a batch size of 6. We employed various backbone networks to comprehensively validate our approach. These include classic single-scale CNN encoders like the ResNet~\cite{he2016deep} series with dilated convolutions, single-scale Transformer encoders like the ViT~\cite{dosovitskiy2020image} series, multi-scale dense prediction network HRNet~\cite{sun2019deep} series, and multi-scale deformable convolution series InternImage~\cite{wang2023internimage}. For models using ViT and ResNet-50 backbones, we utilized the Adam optimizer with a learning rate of $2\times 10^{-5}$ and a weight decay rate of $1\times 10^{-6}$. For models using HRNet backbones, we utilized the SGD optimizer with a learning rate of $0.01$ and a weight decay rate of $5\times 10^{-4}$, and the momentum weight is $0.9$. And for models using InternImage backbones, we utilized the AdamW optimizer with a learning rate of $6\times 10^{-5}$and $2\times 10^{-5}$ for the \textit{base} and the \textit{large} model respectively, and a weight decay rate of $0.05$, the $\beta$ for the optimizers are $(0.9, 0.999)$. A polynomial learning rate decay scheduler was used. 

We maintained alignment with the settings from~\cite{ye2022inverted}. For single-scale image backbones, we first extract the intermediate backbone features at specified depths, then convert their spatial and channel dimension by convolution layers to build a multi-scale feature pyramid. The last three layers were taken as the multi-scale input for BridgeNet. Correspondingly, BFE and TFR were applied only to features from these last three scales, ensuring a favorable trade-off between performance and inference speed. For multi-scale image backbones, they directly produce a feature pyramid which is acceptable for BridgeNet.

For ViT-base and ViT-large, the initial output channel numbers of the preliminary decoders were 768 and 1024, respectively. The number of heads in the multi-head attention of BFE and TPP was set to 2. The downsampling ratios for query vectors were 2, and for key and value vectors, they were $[2, 4, 8]$. Regarding the HRNet series models, following the settings of previous multi-scale multi-task methods~\cite{vandenhende2020mti,bruggemann2021exploring}, we utilized features from all four scales. The downsampling ratios for key and value vectors were $[2, 4, 8, 16]$. For HRNet-w18 and HRNet-w48, the initial output channel numbers of the preliminary decoders were set at 144 and 384, respectively. For the InterImage-B and InterImage-L backbones, the initial output channel numbers of the preliminary decoders were 896 and 1280 respectively. Regarding the scale of TFR, we established three different sizes: \textit{base} with 2 layers, \textit{large} with 4 layers, and \textit{huge} with 6 layers. Each TFR layer followed the standard design of Hybrid Dilated Convolution (HDC)~\cite{wang2018understanding}, employing three layers of depth-wise separable dilated convolutions with dilation rates of $[1, 2, 5]$. This was done to avoid the grid effect brought about by multiple layers of dilated convolutions and to achieve as large a receptive field as possible.

\begin{figure*}[t]
  \centering
  \includegraphics[width=0.98\linewidth]{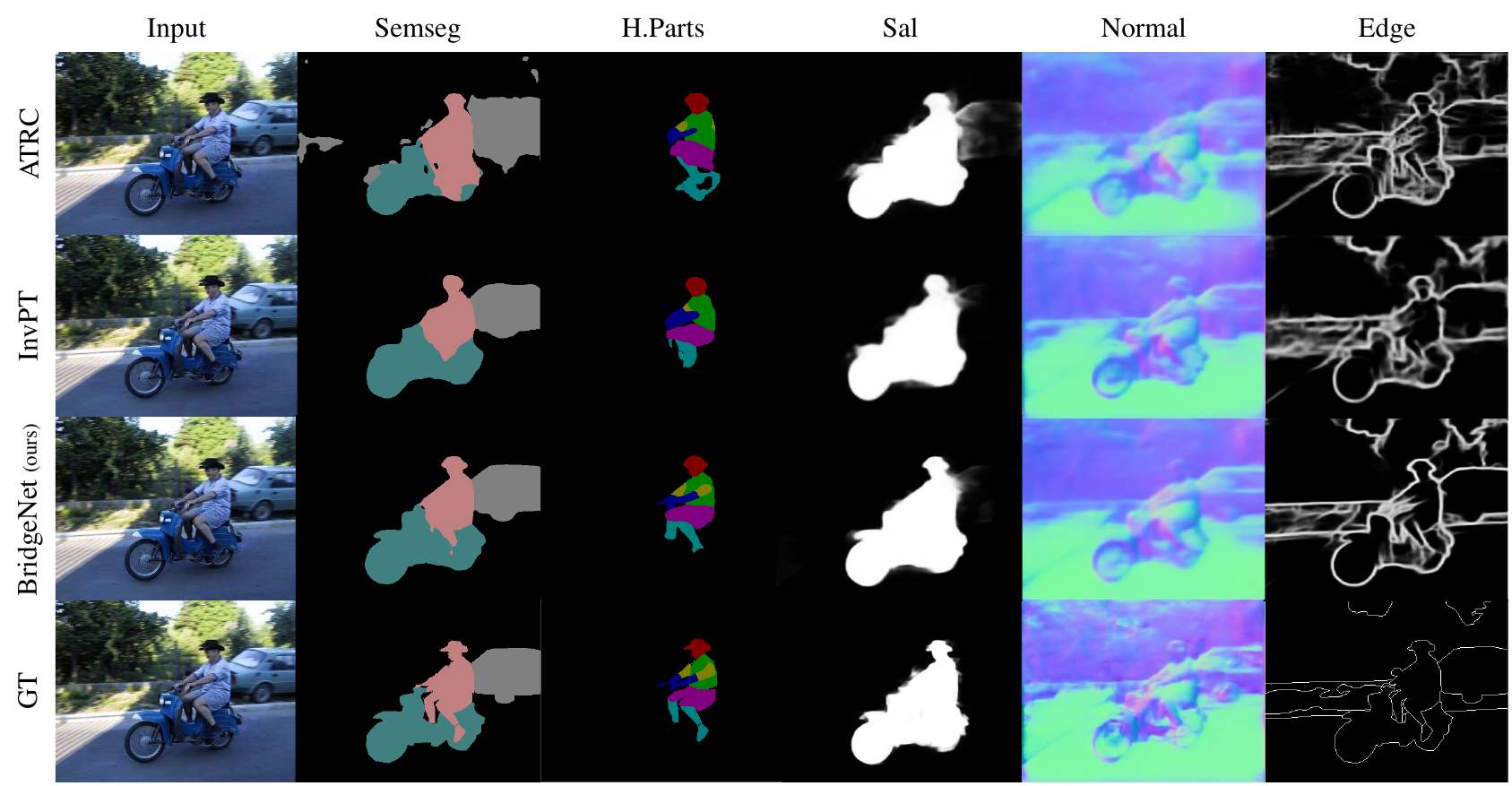}
  \caption{As visualized above, we compare with SOTA works (ATRC\cite{bruggemann2021exploring} and InvPT~\cite{ye2022inverted}) on PASCAL Context 5 tasks. Our BridgeNet clearly produces more precise predictions on each task. We also use the sample example image provided in~\cite{ye2022inverted} paper for comparisons.}
  \label{fig:com_pascal}
\end{figure*}

\noindent\textbf{Baselines:} Following the previous works, we use the same baseline setting by employing a simple multi-task baseline with a shared encoder and multiple decoders. The features from the backbone network's output are directly fed into task-specific decoders or prediction heads, each receiving label supervision from its respective task.

\subsection{Comparison with SOTA Methods}
\subsubsection{Overall Comparisons}
We compared BridgeNet with various state-of-the-art multi-task methods on the NYUD-v2, Cityscapes and PASCAL Context datasets. We employed InternImage-L as the backbone for training on NYUD-v2 and PASCAL Context, and ViT-large as the backbone for Cityscapes. As shown in Table~\ref{tab:compare1}, across all the compared tasks, our method surpasses previous works significantly in all of the tasks. On the NYUD-v2 dataset, the tasks include semantic segmentation (Seg.), depth estimation (Dep.), surface normal estimation (Norm.), and edge detection (Edge.). For the NYUD-v2 comparisons, we used TFR-\textit{huge}, and BridgeNet outperforms previous methods on all tasks. Particularly significant improvements are achieved in semantic segmentation and depth estimation. Compared to InvPT~\cite{ye2022inverted}, our method enhances by $+3.01$ \textit{mIoU} and $-0.0528$ \textit{rmse}, respectively. Compared to the previous state-of-the-art multi-scale interaction MTI-Net~\cite{vandenhende2020mti}, we achieve a significant improvement of $+10.60$ \textit{mIoU} and $-0.0710$ \textit{rmse}. This improvement is substantial. Noticeable enhancements are also seen in the other two tasks. Qualitative comparisons are shown in Fig.~\ref{fig:com}, which further prove that BridgeNet yields better prediction quality.

On the Cityscapes, the tasks include semantic segmentation (Seg.) and depth estimation (Dep.), and we use TFR-\textit{huge}. Compared with encoder-focused methods which only utilize low-level representations in task-generic features for task interactions, our method significantly achieves better performance on both tasks. Also, our BridgeNet surpasses the best performing previous decoder-focused InvPT~\cite{ye2022inverted} by $+1.42$ \textit{mIoU} and $-0.025$ \textit{rmse}. We also make qualitative comparisons in Fig.~\ref{fig:com_citys} to further validate the effectiveness of our method.

\begin{figure*}[t]
  \centering
  \includegraphics[width=0.85\linewidth]{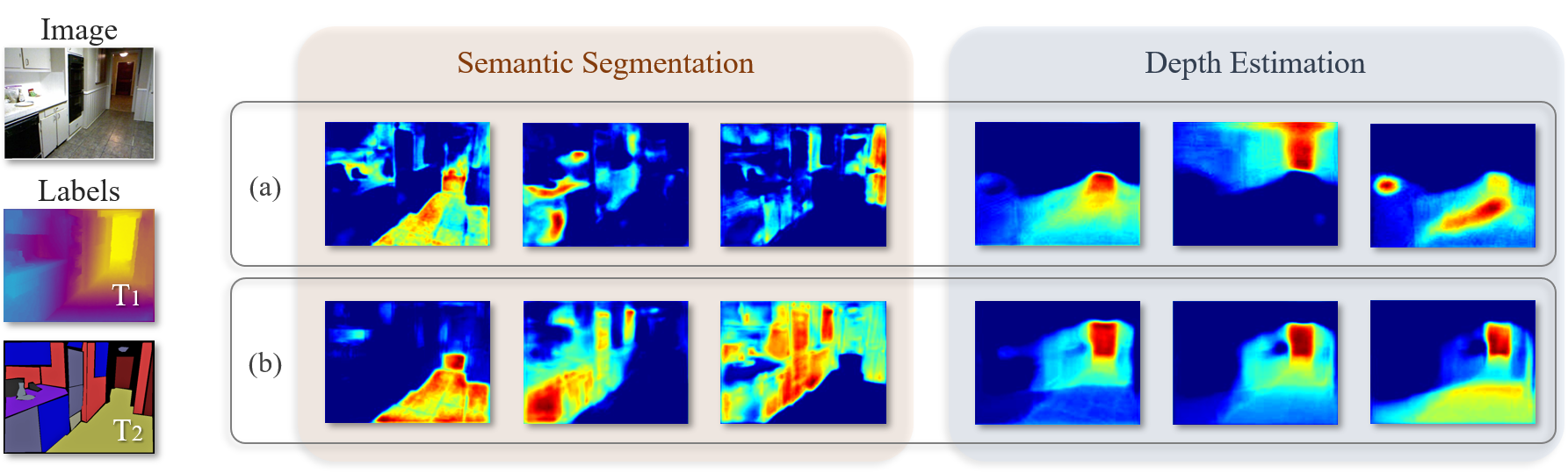}
  \caption{The visualization of feature maps at scale 0 by MTI-Net \textbf{(a)} and our BridgeNet \textbf{(b)}. In semantic segmentation \textbf{(left)}, our method presents more consistency with clear edges for the same semantic region. The brighter area means more attention is paid in those areas. In depth estimation \textbf{(right)}, our method also produces more accurate edges for areas with different depths, such as the door in the faraway distance which is supposed to be the brightest area. $T_{1}$ and $T_{2}$ represent semantic segmentation and depth estimation labels respectively.}
  \label{fig:ab2}
\end{figure*}

On the PASCAL Context dataset, the tasks include semantic segmentation (Seg.), human body part segmentation (H.Parts), surface normal estimation (Norm.), saliency estimation (Sal.), and edge detection (Edge.). We used TFR-\textit{large} for BridgeNet. Due to the increased number of tasks, achieving a balance in performance across multiple tasks becomes challenging, and maintaining superiority across all tasks over previous works is not easy to achieve. Nevertheless, even under these conditions, we achieved balanced promotions in all tasks. More significant improvement occurred in human body part segmentation (H.Parts) and saliency estimation (Sal.), resulting in $+3.72$ \textit{mIoU} and $-0.77$ \textit{mErr}. Besides, we also conduct qualitative comparisons in Fig.~\ref{fig:com_pascal}. Furthermore, we visualized the task-specific features refined through TFR, as shown in Fig.~\ref{fig:ab2}. In comparison to works lacking the TFR structure, such as MTI-Net~\cite{vandenhende2020mti}, the task-specific features generated by our method exhibit clearer high-level semantics, highlighted regions possess sharper boundaries, and overall coherence is improved.

\subsubsection{Comparisons on Different Backbones}
Different from InvPT~\cite{ye2022inverted} which is specially designed for ViT backbones, our method is adaptive to different types of backbones. We also employed different backbones to compare with various established methods, ensuring the effectiveness of our method across diverse conditions. We utilized two different sizes of ViTs~\cite{dosovitskiy2020image}, ViT-base and ViT-large, two different sizes of HRNet~\cite{sun2019deep}, HRNet-w18 and HRNet-w48, as well as the commonly used ResNet-50d with dilated convolutions~\cite{he2016deep,chen2017deeplab}. As observed, our BridgeNet demonstrates robust performance across various types of backbone networks.

When using the HRNet series backbone networks, our method exhibits comparable performance to other methods in the Semseg. task, while showcasing significant improvements in other spatial geometric estimation tasks. On ResNet-50d, though our method achieves competitive performance, the performance among different tasks is more balanced, which means our method does not tend to sacrifice performance on one task to enhance another. When employing larger parameter backbone networks such as the ViT series, our method shows significant enhancements in high-level semantic understanding tasks like Semseg. and H.Parts. The reason behind this phenomenon is that when the backbone network has fewer parameters, its feature extraction capacity might be limited. Simpler geometric estimation tasks (Norm., Edge.) might not have reached saturation, allowing the model to learn useful representations from richer semantic tasks (Semseg.) and thereby enhance the performance. Conversely, when the backbone network has more parameters, it can capture precise image representations, reaching saturation in geometric estimation tasks. At this point, further improvements become challenging. However, larger feature space and richer representations become advantageous for tasks that require advanced information (Semseg. and H.Parts). This phenomenon is also observed in~\cite{ye2022inverted}.

\subsubsection{Comparisons on Model Size and Cost}
To further show the efficiency of our BridgeNet, we compare our method with some of the previous SOTA works with the consideration of computation cost, parameter amount, latency, and memory cost regarding a single image input. We test all models on a single NVIDIA 3090 GPU under the same condition. As shown in Table~\ref{tab:cost}, our method achieves competitive performance with ViT-base compared with other methods applied with ViT-large, and under the same ViT-large backbones, our method shows clear superiority over previous works on the majority of tasks, especially on Seg., Sal. and Norm. Benefits from the high-efficiency interactions brought by BFE, our BridgeNet with the configuration of TFR-\textit{large} reduces over 100 GFLOPs compared with~\cite{xu2018pad,vandenhende2020mti,bruggemann2021exploring}, and 24 GFLOPs compared with InvPT~\cite{ye2022inverted}. Considering the memory cost for single image inference, our BridgeNet significantly surpasses methods with heavy Multi-scale Feature Propagation Module (FPM) like~\cite{vandenhende2020mti,bruggemann2021exploring}, with using less than half of the memory. Though our method is slightly slower when comparing the inference latency with other methods, which is due to the two-step extraction and refinement involving the bridge feature, however the advantages in model size, FLOPs, memory consumption, etc. prove that our BridgeNet saves a lot of additional computational overhead for multi-task interactions. 

For different backbone configurations, our method shows better performance with less computation costs on InternImage backbones. Compared with ViT-base, our method performs better with compatible FLOPs and Params on InternImage-B. When compared with ViT-large, our method achieves better performance on InternImage-L even with fewer parameters (-57 MParams) and less computation costs (-118 GFLOPs).

\begin{table}[t]
\caption{Ablation analysis of the components of BridgeNet. The experiments are conducted on NYUD-v2.}
\centering
 \setlength{\tabcolsep}{1.1mm}{\scalebox{1.0}{
\begin{tabular}{lccccc}
\toprule
\multirow{2}{*}{Models} & Seg. & Dep. & Norm. & Edge. &$\Delta_{M T L}$ \\
&\textit{mIoU} ↑&\textit{rmse} ↓ &\textit{mErr} ↓ &\textit{odsF} ↑&(\%) ↑ \\
\cmidrule(lr){1-1}\cmidrule(lr){2-5}\cmidrule(lr){6-6}
STL	&50.95	&0.5698	&19.08	&78.28	&- \\
MTL basline	&48.30	&0.5605&	19.08&	77.42&	-1.17\\
\cmidrule(lr){1-1}\cmidrule(lr){2-5}\cmidrule(lr){6-6}
$+$BFE	&50.30	&0.5367	&19.00	&77.40	&0.96\\
$+$BFE$+$TPP	&50.22	&0.5312	&18.97	&77.68	&1.29\\
$+$BFE$+$TPP$+$TFR-\textit{huge}	&51.14	&0.5186	&18.92	&77.98	&2.45\\
\cmidrule(lr){1-1}\cmidrule(lr){2-5}\cmidrule(lr){6-6}
$\Delta_{\tau_{j}} / (\%)$ ↑ &0.37	&8.99	&0.84	&-0.38	&-\\
\bottomrule
\end{tabular}
}}
\label{tab:1}
\end{table}

\begin{table}[t]
\caption{Ablation analysis of refine strategies. The experiments are conducted on NYUD-v2.}
\centering
 \setlength{\tabcolsep}{1.2mm}{\scalebox{1.0}{
\begin{tabular}{lcccccc}
\toprule
\multirow{2}{*}{Strategies} & Params& Seg. & Dep. & Norm. & Edge. &$\Delta_{M T L}$ \\
& (M)&\textit{mIoU} ↑&\textit{rmse} ↓ &\textit{mErr} ↓ &\textit{odsF} ↑&(\%) ↑ \\
\cmidrule(lr){1-2}\cmidrule(lr){3-6}\cmidrule(lr){7-7}
STL &-	&50.95	&0.5698	&19.08	&78.28	&- \\
MTL basline&-	&48.30	&0.5605&	19.08&	77.42&	-1.17\\
\cmidrule(lr){1-2}\cmidrule(lr){3-6}\cmidrule(lr){7-7}
Add	&+0 &49.94	&0.5354	&18.89	&77.34	&0.96\\
Concatenate	&+7.3&	50.20&	0.5303&	18.95&	77.10&	1.16\\
Add + InvPT	&+30.7&	50.33&	0.5386&	18.92&	77.32&	0.97\\
TFR-\textit{base}	&+15.9&	50.36&	0.5267&	19.04&	77.86&	1.52\\
TFR-\textit{large}	&+31.7	&50.86	&0.5247	&18.95	&77.84	&1.96\\
TFR-\textit{huge}	&+47.6	&51.14	&0.5186	&18.92	&77.98	&2.45\\
\bottomrule
\end{tabular}
}}
\label{tab:2}
\end{table}

\begin{figure*}[t]
  \centering
  \includegraphics[width=1.0\linewidth]{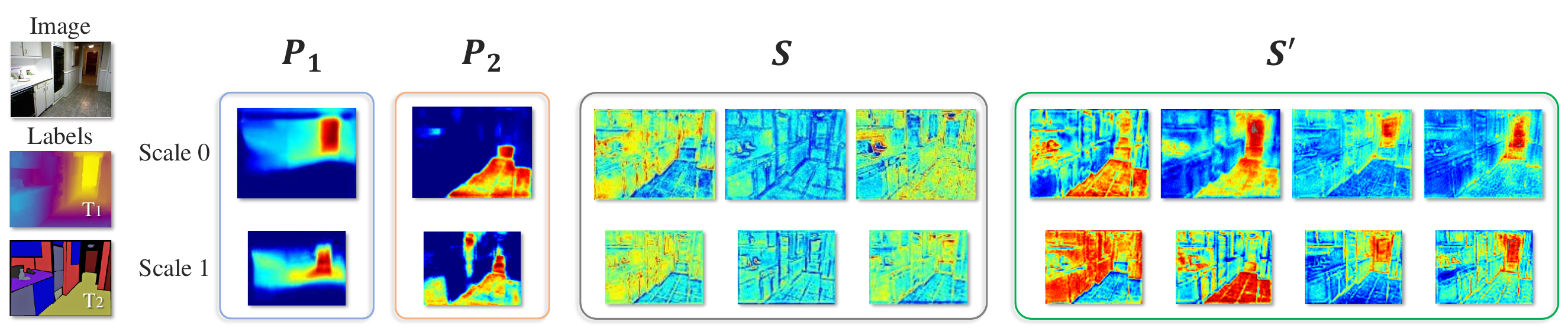}
  \caption{We randomly visualized feature maps from Bridge Features ($\mathbf{S^{\prime}}$), task-generic features ($\mathbf{S}$), and task-specific features ($\mathbf{P}$) at scales 0 and 1. In comparison to $\mathbf{S}$, $\mathbf{S^{\prime}}$ exhibits more high-level representations, highlighted areas on different distances, and semantic objects. Additionally, $\mathbf{S^{\prime}}$ retains the rich low-level features obtained in $\mathbf{S}$, such as edges and textures.}
  \label{fig:cc}
\end{figure*}

\begin{table*}[t]
\caption{Comparison of different task sets on NYUD-v2.}
\centering
\setlength{\tabcolsep}{4.0mm}{\scalebox{1.0}{
\begin{tabular}{lccccccccc}
\toprule
\multirow{2}{*}{Models} &Seg. &$\Delta_{\tau_{seg}}$ &Dep. & $\Delta_{\tau_{dep}}$ &Norm. &$\Delta_{\tau_{norm}}$ &Edge. &$\Delta_{\tau_{edge}}$ &$\Delta_{M T L}$\\
&\textit{mIoU}↑ &(\%) &\textit{rmse}↓ &(\%) &\textit{mErr}↓ &(\%) &\textit{odsF}↑ &(\%) &(\%) \\
\cmidrule(lr){1-1}\cmidrule(lr){2-3}\cmidrule(lr){4-5}\cmidrule(lr){6-7}\cmidrule(lr){8-9}\cmidrule(lr){10-10}
STL	&50.95&	-	&0.5698&	-	&19.08	&-	&78.28	&-	&-\\
\cmidrule(lr){1-1}\cmidrule(lr){2-3}\cmidrule(lr){4-5}\cmidrule(lr){6-7}\cmidrule(lr){8-9}\cmidrule(lr){10-10}
\multirow{6}{*}{BridgeNet (TRF-\textit{base})} & 52.73 &	+3.49&	0.5247&	+7.92	&-	&-	&-&	-	&5.70\\
&50.30	&-1.28	&-	&-	&19.03	&+0.26	&-	&-	&-0.51 \\
&-	&-	&0.5416	&+4.95	&18.90	&+0.94	&-	&-	&2.94 \\
&50.20	&-1.47	&0.5305	&+6.90	&19.03	&+0.26	&-	&-	&1.90 \\
&-& -& 0.5351	&+6.09	&18.89	&+1.00	&77.60	&-0.87	&2.07 \\
& 50.36	&-1.16	&0.5267	&+7.56	&19.04	&+0.21	&77.86	&-0.54 &	1.52\\
 \bottomrule
\end{tabular}
}}
\label{tab:taskset1}
\end{table*}

\begin{table*}[t]
\caption{Comparison of different task sets, we use the TFR-\textit{base} in decoding stage.}
\centering
\setlength{\tabcolsep}{2.52mm}{\scalebox{1.0}{
\begin{tabular}{lcccccccccccccccccccc}
\toprule
 \multirow{2}{*}{Models} & Seg. &$\Delta_{\tau_{seg}}$ &H.Parts &$\Delta_{\tau_{hpart}}$ & Sal. &$\Delta_{\tau_{sal}}$ &Norm. &$\Delta_{\tau_{norm}}$ &Edge.&$\Delta_{\tau_{edge}}$ &$\Delta_{M T L}$\\
&\textit{mIoU}↑ &(\%) &\textit{mIoU}↑ &(\%) &\textit{maxF}↑ &(\%)&\textit{mErr}↓ &(\%) &\textit{odsF}↑ & (\%) &(\%)\\
\cmidrule(lr){1-1}\cmidrule(lr){2-3}\cmidrule(lr){4-5}\cmidrule(lr){6-7}\cmidrule(lr){8-9}\cmidrule(lr){10-11}\cmidrule(lr){12-12}
STL	&79.69	&-	&71.15	&-	&84.77	&-	&13.26	&-	&73.00	&-	&- \\
\cmidrule(lr){1-1}\cmidrule(lr){2-3}\cmidrule(lr){4-5}\cmidrule(lr){6-7}\cmidrule(lr){8-9}\cmidrule(lr){10-11}\cmidrule(lr){12-12}
\multirow{6}{*}{BridgeNet (TRF-\textit{base})} & 80.70	&+1.27	&71.95	&+1.12	&-	&-	&-	&-	&-	&-	&1.20 \\
& 78.93	&-0.95	&-	&-	&85.20	&+0.51	&-	&-	&-	&-	&-0.22 \\
&79.68	&-0.01	&70.74	&-0.58	&85.19	&+0.50	&-	&-	&-	&-	&-0.03 \\
&-	&-	&-	&-	&85.12	&+0.41	&13.35	&-0.68	&2.50	&-0.68&	-0.31 \\
&76.71	&-3.74	&67.33	&-5.37	&84.79	&+0.02	&13.49	&-1.73&	-	&-	&-2.70 \\
&77.98	&-2.15	&68.19	&-4.16	&85.06	&+0.34	&13.48	&-1.65&	72.96	&-0.05	&-1.54 \\
 \bottomrule
\end{tabular}
}}
\label{tab:taskset2}
\end{table*}

\subsection{Ablation Study}
The ablation experiments are divided into three sections. In section 4.3.1, Model Analysis, we delve into the significance of the three core modules that constitute BridgeNet: BFE, TPP, and TFR. We provide comprehensive visual analysis in this regard. In section 4.3.2, Refine Strategies Analysis, we explore how different methods of feature refinement impact the final predictive performance, including varying sizes of TFR and other techniques. In section 4.3.3, Task Set Analysis, we investigate how the model's performance changes when dealing with different combinations of tasks in multi-task learning. For all the experiments, we employ ViT-base as the backbone network.

\subsubsection{Model Analysis} 
These experiments were conducted on the NYUD-v2 dataset. Initially, we trained and tested the performance of single-task learning (STL) on each individual task. Then, we calculated the performance of the multi-task learning baseline (MTL baseline) and progressively incorporated BFE, TPP, and TFR. The comparative results are presented in Table~\ref{tab:1}. As evident, due to the varying number of tasks and their differences, each task pursues a distinct optimization objective. Consequently, MTL baseline exhibits a noticeable performance decline in comparison to STL. Upon introducing BFE over the baseline, performance improvement is observed, nearing the level of advanced models like InvPT~\cite{ye2022inverted} shown in Table~\ref{tab:compare2}. However, at this stage, without TPP's assistance in obtaining more discriminative task-specific features, the quality of the Bridge Features generated by BFE is not optimal. Hence, with the introduction of TPP, further performance enhancement is achieved.

To better harness the bridge features, simple pixel-wise addition of feature maps is insufficient. Therefore, after adding TFR, task-specific features experience superior optimization. From Table 3, it can be observed that the performance improvement is more significant for Dep. among the four tasks. With each of our designed modules incorporated, the model's average multi-task performance $\Delta_{MTL}$ shows improvement. Despite encountering negative transfer among tasks, which prevents surpassing STL performance in Edge., the overall average performance across multiple tasks still improves. This underscores the effectiveness of our model design.

We also conducted a comprehensive visual analysis. As shown in Fig.~\ref{fig:cc}. We selected bridge features ($\mathbf{S^{\prime}}$) and compared them with task-generic features ($\mathbf{S}$) and task-specific features ($\mathbf{P}$) at different scales. These results are presented in Fig.~\ref{fig:cc}. The distribution of task-specific features $\mathbf{P}$ is similar to their corresponding labels and tends to emphasize the relevant regions under supervision. For example, they highlight regions with significant depth variations, as well as distinct semantic areas like floors, walls, and cabinets. This suggests that they possess distinct task-specific perception capabilities. On the other hand, task-generic features $\mathbf{S}$ exhibit rich low-level representations, such as edges and textures. Bridge features $\mathbf{S^{\prime}}$ exhibit both task-specific perception capabilities and low-level representations. This demonstrates that $\mathbf{S^{\prime}}$ has the potential to serve as an intermediate medium for multi-task interaction, ensuring the integrity of feature interactions.

\subsubsection{Refine Strategies Analysis} 
As discussed in Section 3.4, there are several ways to perform task refinement. Here, we compare the performance of different refinement methods and Task Feature Refiners (TFR) of varying sizes. The results are displayed in Table 4. Simply pixel-wise addition of task-specific features and bridge features, or concatenating and processing them with a single convolutional layer, yields sub-optimal results. Applying InvPT~\cite{ye2022inverted} for global spatial and task modeling on the added features also does not significantly improve performance. As analyzed in Section 3.4, bridge features need to effectively transfer learned representations to task-specific features, where local correlations are more critical, rendering global modeling unnecessary and even introducing redundant parameters and computations. Our TFR-\textit{base}, with only half the parameter count, outperforms the effect of refinement using InvPT ($\Delta_{MTL} 1.52\% > 0.97\%$). All three sizes of TFR consistently achieve improvements among all tasks, effectively striking a balance between performance and computational cost.

\subsubsection{Task Set Analysis}
The intrinsic characteristics of tasks in MTL significantly influences the effectiveness of collaborating. Tasks with close relationships might mutually enhance performance, while tasks with substantial differences could lead to negative transfer and performance degradation. In the domain of multi-task dense prediction, there is a relatively limited number of studies focusing on task composition through ablation experiments. Here, we conducted comprehensive experiments on NYUD-v2 and PASCAL Context datasets to analyze task composition effects.

Table~\ref{tab:taskset1} illustrates different combinations of tasks on the NYUD-v2 dataset. We considered 6 distinct task sets, including cases with 2, 3, and 4 tasks. It can be observed that when fewer tasks are involved, the occurrence of mutual enhancement between tasks becomes more likely. Notably, the most significant mutual enhancement happens between the Seg. and Dep. tasks (Seg. $+3.49\%$ $\Delta_{\tau_{seg}}$, Dep. $+7.92\%$ $\Delta_{\tau_{dep}}$), resulting in an average performance improvement of $+5.70\% \Delta_{MTL}$. This aligns with findings from several studies, indicating a strong latent correlation between depth estimation and semantic segmentation. However, the correlation between the Seg. and Norm. tasks are relatively weak, leading to considerable negative transfer effects (Seg. $-1.28\%$ $\Delta_{\tau_{seg}}$, Norm. $+0.26\%$ $\Delta_{\tau_{norm}}$), resulting in an average performance decline of $-0.51\% \Delta_{MTL}$. The surface normals are aligned with the gradients of depth in the space, establishing a strong correlation between these two tasks, facilitating mutual enhancement (Dep. $+4.95\%$ $\Delta_{\tau_{dep}}$, Norm. $+0.94\%$ $\Delta_{\tau_{norm}}$), and achieving an average performance improvement of $+2.94\% \Delta_{MTL}$. As the number of tasks increases, due to inherent inconsistencies between various task pairs, such as Seg. and Norm., performance tends to decrease. For example, the combination of Seg., Dep., and Norm. results in $+1.90\% \Delta_{MTL}$, with a more noticeable decline in Seg.'s performance. In contrast, the combination of Dep., Norm., and Edge. exhibits good performance, with an average improvement of $+2.07\% \Delta_{MTL}$, as all three tasks are designed to learn spatial geometry and lack evident conflicts.

Overall, among the four tasks, Dep. displays the most pronounced improvement due to its minimal conflicts with the other three tasks, allowing it to extract sufficient information from other tasks. In contrast, Seg. and Edge. exhibit relatively modest enhancements. Seg.'s potential conflicts hinder its enhancement, while Edge.'s phenomenon aligns with the analysis in ATRC~\cite{bruggemann2021exploring}, where it benefits less from the other tasks.

Table~\ref{tab:taskset2} depicts different combinations of tasks on the PASCAL Context dataset. We also examined 6 different task combinations, encompassing 2, 3, 4, and 5 tasks. Among these, the Seg., H.Parts, and Sal. tasks exhibit higher correlation. Learning these tasks together or in pairs does not yield noticeable negative effects. Seg. and H.Parts are tasks involving segmentation at different granularities, and they share a substantial amount of information. This direct information sharing explicitly enhances prediction accuracy of the \textit{person} category in Seg., boosting the Intersection over Union (\textit{IoU}) from $86.10$ to $86.67$. The interaction of high-level semantic information across different granularities benefits both tasks, resulting in an average performance improvement of $+1.20\% \Delta_{MTL}$. Similarly, the Seg. and Sal. tasks share some foreground and background information, leading to no substantial negative impact, with an average performance change of $-0.22\% \Delta_{MTL}$. Learning all three tasks together results in an average performance change of $-0.33\% \Delta_{MTL}$, which is an acceptable compromise. However, introducing Norm. in addition to these three tasks significantly reduces the gains, with an average performance decline of $-2.70\% \Delta_{MTL}$. This observation aligns with the findings on the NYUD-v2 dataset. Furthermore, the addition of Edge. results in a performance rebound, enhancing the relative gains for nearly every task. This underscores the importance of Edge. for most dense prediction tasks. When learning Sal., Norm., and Edge., the three relatively simpler tasks together, the average performance is $+0.31\% \Delta_{MTL}$. 

\section{Conclusion}
In this paper, we discovered the limitations of previous multi-task dense prediction methods in terms of incomplete levels of representations, less discriminative semantics in feature participants, and inefficient pair-wise task interaction processes. To address these challenges, we introduced a novel framework called BridgeNet. Specifically, our method adopts a shared encoder backbone to generate task-generic features. In the early decoding stage, a preliminary decoder with Task Pattern Propagation (TPP) is used to generate high-quality task-specific features. We introduced the Bridge Feature Extractor (BFE) to interact between globally queried task-specific features and backbone task-generic features, selecting task-specific perception information and generating multi-scale bridge features. The Task Feature Refiner (TFR) injects the representation of bridge features and iteratively optimizes the final task predictions, resulting in the ultimate task predictions. Compared to previous structures, our proposed method ensures the integrity of feature interaction, striking a good balance between performance and cost.

In our experiments, we evaluated our method on the widely used multi-task learning benchmark datasets, NYUD-v2, Cityscapes and PASCAL Context. By comparing with Single Task Learning (STL), Multi-Task Learning Baseline (MTL baseline), PAD-Net, MTI-Net, InvPT, and our BridgeNet, we observed the superior performance of BridgeNet across multiple tasks. Furthermore, we conducted a comprehensive visual analysis to demonstrate the advantage of bridge features over the other two types (task-generic and task-specific features). Compared to models generating and utilizing task-generic and task-specific features, such as BMTAS and InvPT, BridgeNet dives deeper into feature mining from important dense prediction priors, driving significant performance improvements. We analyzed the reasons for our method's superiority at the feature level. Moreover, through ablation experiments on model components, we analyzed the effectiveness of each component and demonstrated the improvements of our method tailored for dense prediction tasks. Additionally, our ablation experiments on task sets started from the nature of each task to analyze task relationships and studied the impact of different task combinations on multi-task learning performance. By conducting these ablation experiments on task sets, we gained a better understanding of the interactions between tasks and provided guidance for constructing effective multi-task learning models.

In summary, our research addressed the limitations of previous multi-task dense prediction methods through the introduction of the BridgeNet approach. Experimental results show that BridgeNet achieves significant performance improvements in multi-task learning, which were further validated through visual analysis and ablation experiments. Our work provides valuable insights and guidance for research and practice in the field of multi-task dense prediction. Future research can further explore the application and optimization of the BridgeNet method to meet the broader demands of dense prediction tasks.

{\small
\bibliographystyle{ieee_fullname}
\bibliography{egbib}
}

%








\begin{IEEEbiography}[{\includegraphics[width=1in,height=1.25in,clip,keepaspectratio]{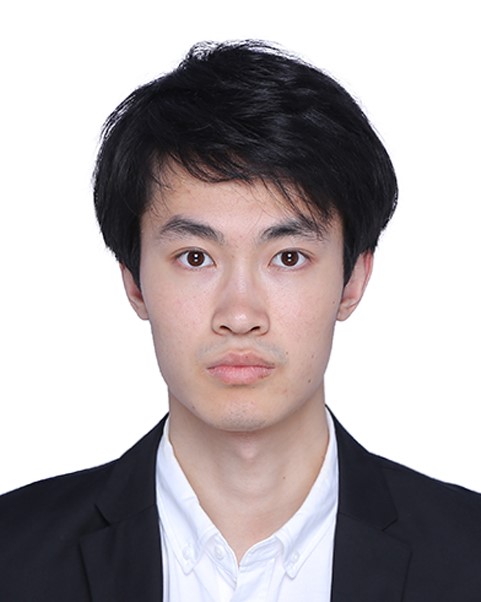}}]{Jingdong Zhang}
is a second-year Ph.D. student in Computer Science and Engineering at Texas A\&M University, specializing in computer graphics and vision. He received his bachelor’s degree in Intelligent Science and Technology from Fudan University. His research interests include multi-task learning, dense scene parsing, neural rendering, 3D reconstruction and generation. He serves as a reviewer for various conferences, including CVPR, ECCV, ICRA, and PG.
\end{IEEEbiography}

\begin{IEEEbiography}[{\includegraphics[width=1in,height=1.25in,clip,keepaspectratio]{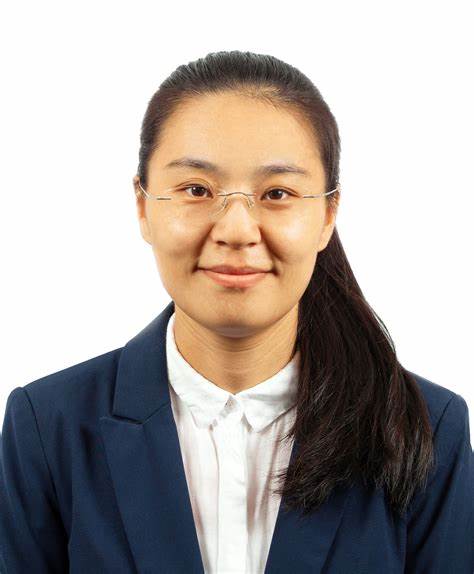}}]{Jiayuan Fan}
received the Ph.D. degree in information engineering from Nanyang Technological University, Singapore, in 2015. After her graduation, she worked as a Research Scientist at the Institute for Infocomm Research, A*STAR, Singapore. She is currently an associate professor with Academy for Engineering and Technology in Fudan University, Shanghai, China. Her main research interests include computer vision, and image forensic analysis and application.
\end{IEEEbiography}

\begin{IEEEbiography}[{\includegraphics[width=1in,height=1.25in,clip,keepaspectratio]{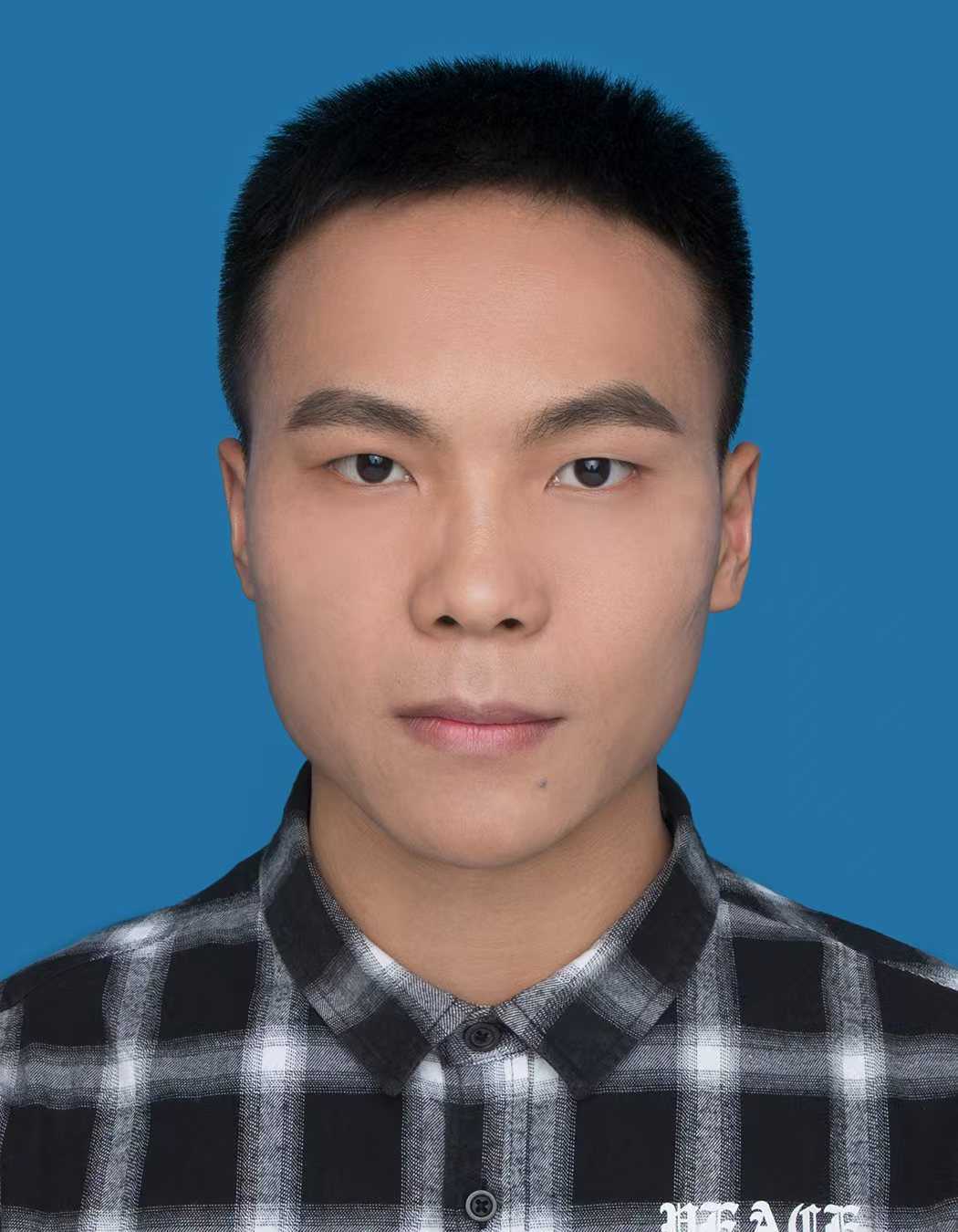}}]{Peng Ye}
is currently pursuing the Ph.D. degree with Fudan University, Shanghai, China. He has published papers in leading journals and conferences, including PAMI, IJCV, CVPR Oral, NeurIPS, ACM MM Oral, ICME Best Student Paper, TGRS, TCSVT, and ICASSP Oral. His research interests include computer vision, network design, and network optimization. He serves as a Reviewer for various journals and conferences, including PAMI, IJCV, TCSVT, CVPR, ECCV, ICCV, and ICLR.
\end{IEEEbiography}

\begin{IEEEbiography}[{\includegraphics[width=1in,height=1.25in,clip,keepaspectratio]{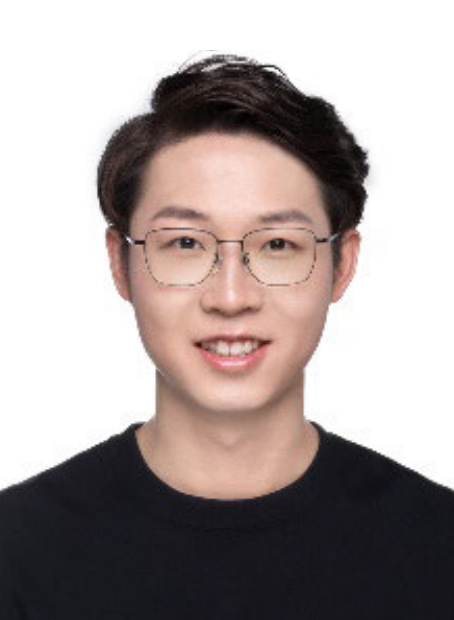}}]{Bo Zhang}
is currently a Research Scientist at the Shanghai Artificial Intelligence Laboratory. He has published 22 papers in top-tier international conferences and journals such as CVPR, NeurIPS, ICML, ICLR, etc. Professionally, he led the development of the 3DTrans open-source project, which won the championship of the Waymo International Challenge and accumulated over 5k stars. Additionally, he is committed to promoting the rapid application of multi-modal large language models in various scenarios, such as scientific document understanding, scientific research surveys, mathematical reasoning, and autonomous driving.
\end{IEEEbiography}

\begin{IEEEbiography}[{\includegraphics[width=1in,height=1.25in,clip,keepaspectratio]{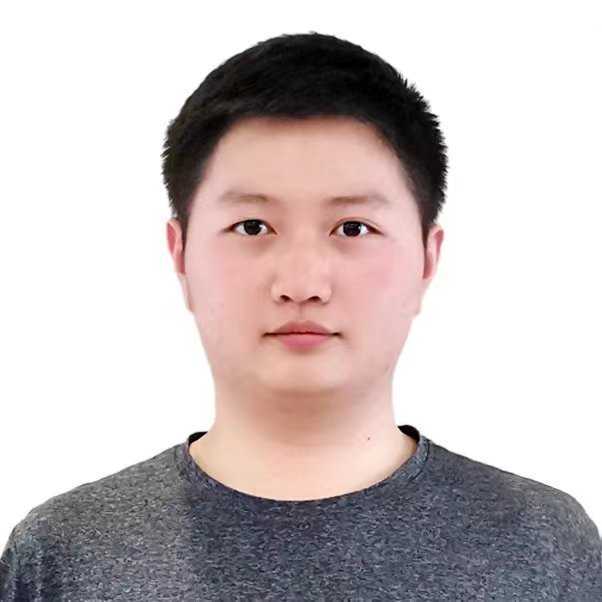}}]{Hancheng Ye}
received his B.S. and M.S. degrees in Electronic Engineering at the School of Information Science and Technology from Fudan University. He is currently pursuing a Ph.D. degree at Duke University. His primary research interests focus on efficient machine learning, model compression, and multimodal learning.
\end{IEEEbiography}

\begin{IEEEbiography}[{\includegraphics[width=1in,height=1.25in,clip,keepaspectratio]{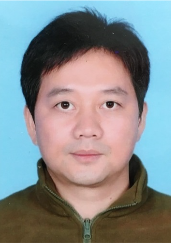}}]{Baopu Li}
obtained his Ph.D degree in the field of computer vision and robotics from the
Chinese University of Hong Kong in the year 2008. After that, he was in the academic field for another 8 years. Since 2016, he changed his career path to industry in the USA. His major research and development interests include Auto machine learning (ML), low-level image processing, video understanding and so on together with their applications on cloud and edge devices.
\end{IEEEbiography}

\begin{IEEEbiography}[{\includegraphics[width=1in,height=1.25in,clip,keepaspectratio]{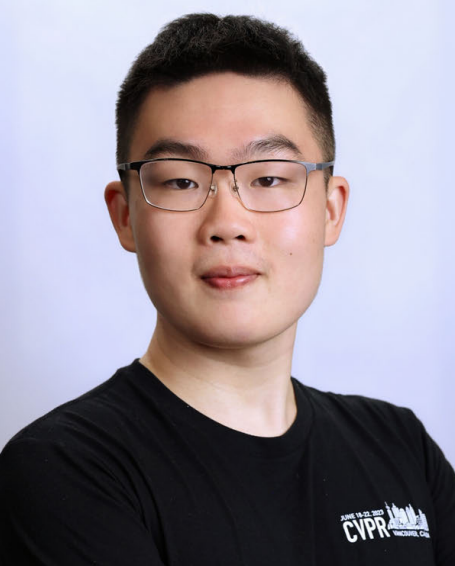}}]{Yancheng Cai}
is a second-year PhD student in Computer Science and Technology at the University of Cambridge, specializing in artificial intelligence, computer vision, computer graphics, and robotics. He completed his undergraduate studies at Fudan University in the Excellence Class of Intelligent Science and Technology, ranking first in his engineering department for four consecutive years. He has received two National Scholarships, the Fudan University Top Ten Students award, numerous individual awards for academic and research excellence, as well as the College Star and Shanghai Outstanding Graduate titles. Formerly a Research Assistant with the Fudan’s Embedded Deep Learning Laboratory, he continued as a Research Assistant with Stanford’s Computer Vision Laboratory in 2022. As the first author, he has published academic papers in CVPR 2023, SIGGRAPH Asia 2024, T-IP, and Neurocomputing. Additionally, he has served as a reviewer for top conferences such as CVPR 2022, CVPR 2023, ECCV 2022, and ECCV 2024, and as a reviewer for T-PAMI. For more information visit the link (https://caiyancheng.github.io/academic.html).
\end{IEEEbiography}

\begin{IEEEbiography}[{\includegraphics[width=1in,height=1.25in,clip,keepaspectratio]{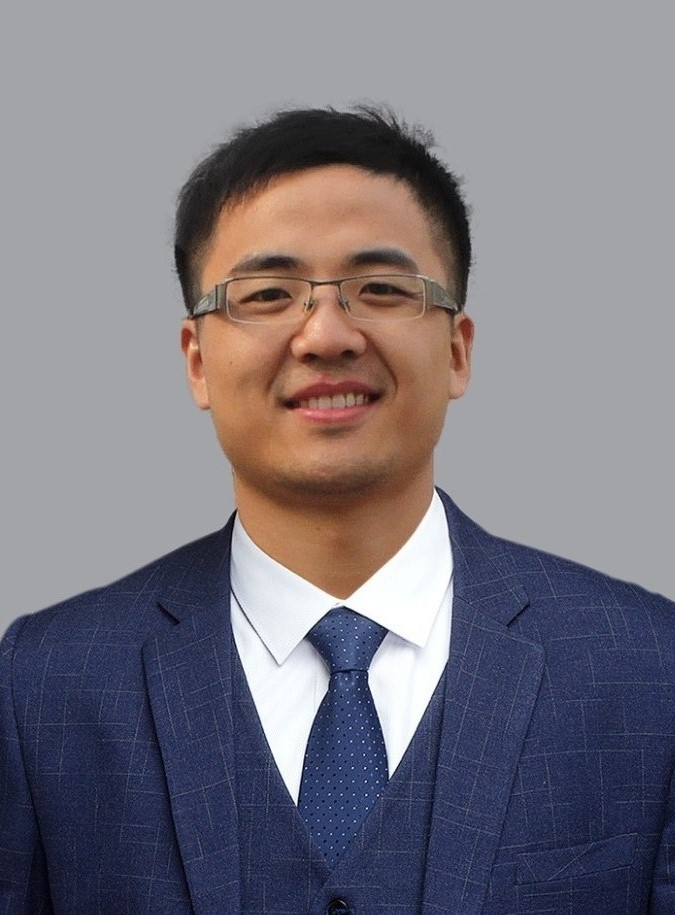}}]{Tao Chen}
(Senior Member, IEEE) received the Ph.D. degree in information engineering
from Nanyang Technological University, Singapore, in 2013. He was a Research Scientist with the Institute for Infocomm Research, A*STAR, Singapore, from 2013 to 2017, and a Senior Scientist with the Huawei Singapore Research Center
from 2017 to 2018. He is currently a Professor with the School of Information Science and Technology, Fudan University, Shanghai, China. His main research interests include computer vision and machine learning.
\end{IEEEbiography}

\end{document}